%% file: main.tex
\documentclass[journal,twoside,web]{ieeecolor}

\input{preamble}

\def\BibTeX{{\rm B\kern-.05em{\sc i\kern-.025em b}\kern-.08em
    T\kern-.1667em\lower.7ex\hbox{E}\kern-.125emX}}
\begin{document}

\title{Facial Spatiotemporal Graphs: \\ 
Leveraging the 3D Facial Surface \\ 
for Remote Physiological Measurement}

\author{
    Sam Cantrill,
    David Ahmedt-Aristizabal, 
    Lars Petersson,
    Hanna Suominen,
    Mohammad Ali Armin
    \thanks{
        This work was supported by the MRFF Rapid Applied Research Translation grant (RARUR000158), CSIRO AI4M Minimising Antimicrobial Resistance Mission, and Australian Government Training Research Program (AGRTP) Scholarship. 
    }
    \thanks{
        Sam Cantrill, David Ahmedt-Aristizabal, Lars Petersson, and Mohammad Ali Armin are with the Imaging and Computer Vision Group, Data61, CSIRO, Canberra, Australia (e-mail: sam.cantrill@data61.csiro.au; david.ahmedtaristizabal@data61.csiro.au; lars.petersson@data61.csiro.au; ali.armin@data61.csiro.au).
    }
    \thanks{
        Sam Cantrill and Hanna Suominen are with the School of Computing, College of Systems and Society, Australian National University, Canberra, Australia (e-mail: sam.cantrill@anu.edu.au; hanna.suominen@anu.edu.au).
    }
    \thanks{
        Hanna Suominen is also with the School of Medicine and Psychology, College of Science and Medicine, Australian National University, Canberra, Australia (e-mail: hanna.suominen@anu.edu.au).
    }
    \thanks{
        Hanna Suominen is also with the University of Turku, Turku, Finland (e-mail: hajasu@utu.fi).
    }
}

\maketitle

\input{sec/0_abstract}


\input{sec/1_intro}
\input{sec/2_related}
\input{sec/3_method}
\input{sec/4_experiments}
\input{sec/5_discussion}

\input{sec/6_conclusion}

\section*{Compliance with Ethical Standards}
This study was performed in line with the principles of the Declaration of Helsinki. The experimental procedures involving human subjects described in this paper were approved by CSIRO Health and Medical Human Research Ethics Committee (CHMHREC) [ethics protocol 2022 016 LR] and the Australian National University Human Research Ethics Committee (ANU
HREC) [ethics protocols 2023/403 and 2023/483].

% \clearpage
\section*{References}
\bibliographystyle{IEEEtran}
\vspace{-15pt}
{\footnotesize
\bibliography{main}
}

\end{document}

%% file: preamble.tex
\usepackage{generic}
\usepackage{cite}
\usepackage{amsmath,amssymb,amsfonts}
\usepackage[x11names,table,dvipsnames]{xcolor}
\usepackage{graphicx}
\usepackage{algorithm}
\usepackage{algpseudocode}
\usepackage{hyperref}
\hypersetup{hidelinks}
\usepackage{textcomp}

\usepackage{cleveref}
\usepackage{tabularx}
\usepackage{multirow}
\usepackage{multicol}
\usepackage{amsmath}
\usepackage{amsfonts}
\usepackage{amssymb}
\usepackage{tikz}
\usepackage{pgfplots}
\usepackage{tikz}

\usetikzlibrary{calc}
\usetikzlibrary{shapes.geometric, arrows}
\usetikzlibrary{shapes.arrows}
\usetikzlibrary{positioning}
\usetikzlibrary{backgrounds}
\usepackage{float}
\usepackage{algpseudocode}
\usepackage{booktabs}
\usepackage{algorithm}
\usepackage{pifont}
\usepackage{pgfplots}
\pgfplotsset{compat=1.18}
\usepackage{pgfplotstable}
\usepackage{siunitx} % optional, for units/formatting
\usepgfplotslibrary{groupplots}
\usepgfplotslibrary{colormaps} % for viridis

\newcommand{\cmark}{\ding{51}}%
\newcommand{\xmark}{\ding{55}}%

\usepackage{graphicx} % in preamble
\usepackage{makecell} % preamble

\definecolor{dark_red}{RGB}{180,10,10}
\definecolor{dark_green}{RGB}{10,180,10}
\definecolor{dark_blue}{RGB}{10,10,180}
\definecolor{Maroon}{rgb}{0.75, 0, 0}

\pgfplotsset{compat=1.18}

\usepackage{pifont}

%% file: sec/0_abstract.tex
\begin{abstract}
    Facial remote photoplethysmography (rPPG) methods estimate physiological signals by modeling subtle color changes on the 3D facial surface over time.
    However, existing methods fail to explicitly align their receptive fields with the 3D facial surface—the spatial support of the rPPG signal.
    To address this, we propose the Facial Spatiotemporal Graph (STGraph), a novel representation that encodes facial color and structure using 3D facial mesh sequences-enabling surface-aligned spatiotemporal processing.
    We introduce MeshPhys, a lightweight spatiotemporal graph convolutional network that operates on the STGraph to estimate physiological signals.
    Across four benchmark datasets, MeshPhys achieves state-of-the-art or competitive performance in both intra- and cross-dataset settings.
    Ablation studies show that constraining the model’s receptive field to the facial surface acts as a strong structural prior, and that surface-aligned, 3D-aware node features are critical for robustly encoding facial surface color.
    Together, the STGraph and MeshPhys constitute a novel, principled modeling paradigm for facial rPPG, enabling robust, interpretable, and generalizable estimation.
    Code is available at \href{https://samcantrill.github.io/facial-stgraph-rppg/}{Facial STGraphs}.
\end{abstract}

%% file: sec/1_intro.tex
\input{figures/summary-diagram}

\section{Introduction}
\label{sec:introduction}
\IEEEPARstart{S}{ubtle} changes in skin color occur as blood volume pulses across the facial surface, manifesting a spatiotemporal pattern encoded in facial video~\cite{takano2007heartrate}.
To estimate this signal, models must capture and extract these %subtle color 
changes on the facial surface~\cite{spetlik2018visual,yu2019physnet,yu2022physformer} while remaining robust to confounding factors such as motion and illumination~\cite{gudi2020real,paruchuri2024motion,yuting2025greip}.
To aid this, existing methods often introduce spatial priors to guide the model toward skin regions and suppress irrelevant variation.

Video-based rPPG methods apply facial cropping~\cite{gideon2021waytomyheart,yu2022physformer,yu2023physformerpp,speth2023noncontrast,zou2024rhythmmamba,zou2024rhythmmamba}, masking~\cite{cantrill2024orientation,comas2025beatformer}, or attention~\cite{chen2018deepphys,liu2020multi,liu2023efficientphys} to focus on signal-relevant areas, but they still operate on the 2D image plane, so alignment with the 3D facial surface must be learned implicitly.
STMap-based methods aim to stabilize the input by aggregating color dynamics from predefined regions—using image-plane grids~\cite{niu2018synrhythm,niu2019rhythmnet,lu2021dual,lu2023nest,liu2024rppg} or landmark-based patches~\cite{niu2020video,das2021bvpnet,qian2024dual}—yet these region definitions remain 2D surrogates of the facial surface and lack explicit 3D awareness under pose and occlusion.
Moreover, STMaps flatten region-wise signals into a 2D matrix, discarding spatial relationships between regions and limiting the learning of localized, surface-aligned spatiotemporal patterns (Fig.~\ref{fig:summary-diagram}).
These limitations reveal a fundamental representational mismatch: current inputs are only indirectly aligned with the 3D facial surface-the true support of the signal.
This motivates the question: \textit{Can we directly employ the 3D facial surface as the network’s operand?}

To address this, we formulate facial rPPG as a learning problem defined directly on the 3D facial surface.
We introduce the \emph{Facial Spatiotemporal Graph (STGraph)}, a surface-aligned input representation that encodes localized facial appearance and surface topology over time on sequences of 3D facial meshes.
On top of the STGraph, we propose \emph{MeshPhys}, a lightweight spatiotemporal graph convolutional network that estimates facial rPPG signals directly from this surface-based representation.
By constraining processing to the facial surface and respecting its connectivity, STGraph and MeshPhys aim to reduce motion-induced misalignment, preserve local structure, and leverage the surface itself as a structural prior for rPPG estimation.
The main contributions of this work are summarized as follows:
\begin{itemize}
    \item[i)] We introduce the \textbf{STGraph}, a surface-aligned input representation for facial rPPG that directly encodes localized appearance and the topology of the 3D facial surface over time.

    \item[ii)] We propose \textbf{MeshPhys}, a spatiotemporal graph convolutional network designed to operate on the STGraph, combining surface-aligned spatial modeling with efficient temporal processing to estimate facial rPPG signals.

    \item[iii)] We empirically demonstrate that constraining the model’s receptive field to the 3D facial surface provides a powerful structural prior for rPPG estimation, yielding state-of-the-art or competitive performance across four benchmark datasets with fewer parameters and improved cross-dataset generalization compared to existing video- and STMap-based methods.
\end{itemize}

%% file: figures/summary-diagram.tex
\begin{figure}
    \centering
    \small
    \begin{tikzpicture}[node distance=0.50cm and 0.50cm]
        \node[anchor=south west,inner sep=0] at (0,0){\includegraphics[width=1.00\linewidth]{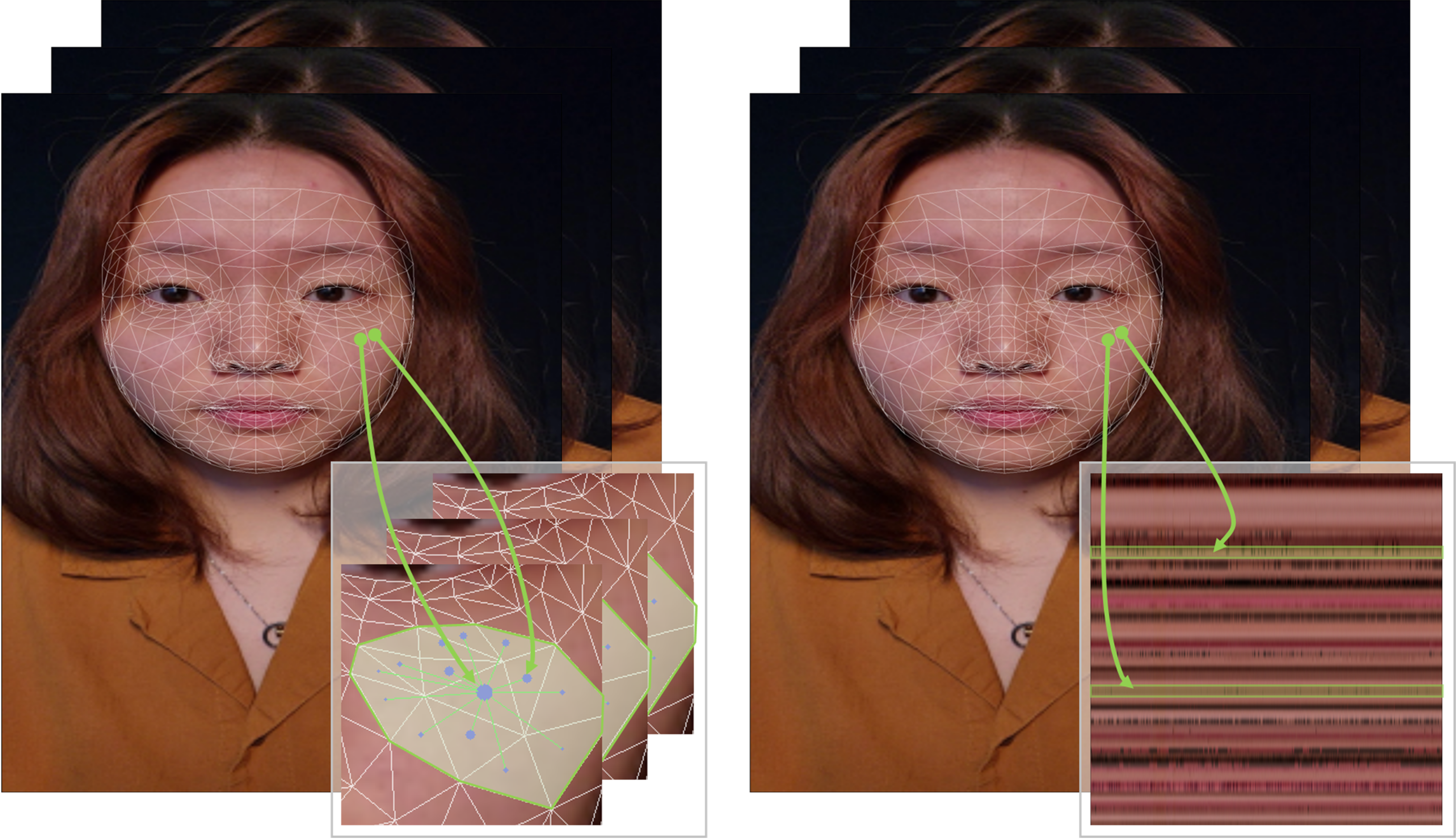}};
        
    \end{tikzpicture}
    \caption{The Facial STGraph (left) preserves facial structure through an adjacency matrix derived from the facial mesh topology-enforcing local spatiotemporal modeling.
    STMaps (right) flatten regions into a 2D matrix neglecting relationships between regions—adjacent entries may be distant on the face-limiting localized modeling.
    }
    \label{fig:summary-diagram}

    \vspace{-1em}
\end{figure}

%% file: sec/2_related.tex
\section{Related Work on Estimating Facial \normalfont{r}PPG}
\label{sec:relatedwork}

\noindent\textbf{Region-based signal-processing methods} 
estimate the rPPG signal by averaging color traces over predefined spatial regions and applying de-mixing~\cite{poh2010advancements,de2013robust,wang2016algorithmic,wang2018singleelement,tulyakov2016samc,pilz2018lgi}. 
These methods often assume minimal motion and lack mechanisms to account for surface correspondence over time. %, making them fragile under head movement, deformation, and illumination shifts.
Consequently, these methods benefit from spatial priors such as facial cropping~\cite{liu2023rppgtoolboxdeepremoteppg}, region selection~\cite{kim2021assessment}, and skin segmentation~\cite{bobbia2019ubfc}, which suppress background and non-skin variation.
Multi-region extensions~\cite{bobbia2019ubfc,kim2021assessment,casado2023face2ppg} improve robustness by aggregating signals across spatial patches, but lack the ability to model complex spatiotemporal dynamics across regions.
These limitations have motivated a shift toward learning-based models that operate directly on richer spatiotemporal representations.

\noindent\textbf{Video-based deep-learning methods} 
estimate physiological signals directly from sequences of raw image frames using spatiotemporal operators~\cite{chen2018deepphys,spetlik2018visual,yu2019physnet,yu2019remote}. 
However, the rPPG signal in video is typically weak and easily entangled with head motion, illumination shifts, background variation, and non-skin regions. 
To address this, many methods introduce priors that constrain the spatial content available to the model. 
Explicit preprocessing steps—such as facial cropping~\cite{gideon2021waytomyheart,yu2022physformer,yu2023physformerpp,speth2023noncontrast,yang2023simper,liu2023rppgtoolboxdeepremoteppg,zou2024rhythmformer,zou2024rhythmmamba}, facial texture mapping~\cite{cantrill2024orientation}, and skin segmentation~\cite{comas2025beatformer}—restrict model input to skin-dominant regions and help suppress background signals and motion artifacts. 
Facial cropping, in particular, is widely adopted across state-of-the-art methods, reflecting a shared reliance on spatial priors for robustness.
However, it does not guarantee consistent surface sampling under out-of-plane rotation or non-rigid deformations.
Meanwhile, implicit priors-such as spatial attention~\cite{chen2018deepphys,liu2020multi,liu2023efficientphys}, spatial feature refinement~\cite{yue2021multimodal,lokendra2022andrppg,shao2023hyperbolic,xiong2024graphphys,stgnet2024xiong,zhao2024toward,stphys2024}, or motion-based regularization~\cite{li2023learning,paruchuri2024motion,speth2024sincp}-aim to encourage models to emphasize regions likely to carry rPPG information. 
However, the receptive field of the model inherently corresponds to different surface regions over time, and the model must implicitly learn to align them.

\noindent\textbf{STMap-based deep-learning methods} 
aim to improve the input by introducing structured representations that encode localized facial color dynamics over time~\cite{niu2018synrhythm}. 
These methods construct a spatiotemporal map (STMap) by sampling the color of predefined coarse spatial regions across video frames. 
There are two main classes of region definitions, each embedding different spatial assumptions.
Image-plane regions are defined as fixed or bounding-box-aligned areas in the video frames—typically using uniform grids~\cite{niu2018synrhythm,niu2019rhythmnet,lu2021dual,lu2023nest,liu2024rppg,condiff2024} anchored to the bounding box.
These assume spatial coherence within the image plane, but actual facial coverage can vary significantly due to out-of-plane rotation and facial motion, leading to motion-confounded features.
Facial-surface regions are typically constructed using patches derived from 2D facial landmarks~\cite{niu2020video,das2021bvpnet,qian2024dual}, which allow for more consistent tracking of facial areas across frames.
However, without 3D awareness, these regions may capture occluded or distorted surface regions during head motion—injecting pose-dependent noise that the model must learn to ignore.
Furthermore, all STMap constructions flatten region-wise features into a 2D matrix, discarding the spatial relationships between regions.
As a result, adjacent entries in the STMap may correspond to distant locations on the facial surface, hindering the model from learning localized spatiotemporal patterns through Euclidean convolution operators.

%% file: sec/3_method.tex
\section{Method}
\label{sec:method}
Section~\ref{sec:method-stgraph} introduces the STGraph, including the definition and construction of its nodes and edges.
Section~\ref{sec:method-MeshPhys} presents MeshPhys, a baseline model designed to operate on the STGraph.
Together, these contributions enable modeling of facial surface appearance dynamics.

\input{sec/3a_stgraph}
\input{sec/3b_meshphys}
\input{sec/3c_losses}

%% file: sec/3a_stgraph.tex
\input{figures/stgraph}

\subsection{Facial Spatiotemporal Graph}
\label{sec:method-stgraph}
We encode the facial surface over a video $I \in \mathbb{R}^{T \times H \times W \times C}$ using a time-indexed $t \in \mathcal{T}$ sequence of spatial graphs that share a common topology.
The STGraph is defined as
\begin{equation}
    \mathcal{G}_{f} = (\mathcal{N}, \mathcal{E}, \mathbf{X}, \mathbf{A}),
    \label{eq:stgraph_def}
\end{equation}
where $\mathcal{N}$ and $\mathcal{E}$ are the node and edge sets, $\mathbf{X} \in \mathbb{R}^{C \times |\mathcal{T}| \times |\mathcal{N}|}$ and $\mathbf{A} \in \mathbb{R}^{|\mathcal{T}| \times |\mathcal{N}| \times |\mathcal{N}| \times E}$ are the node and edge feature tensors.
Each node $n_{t,i} \in \mathcal{N}$ represents the $i$-th facial surface region at time step $t \in \mathcal{T}$ in frame $I_{t}$.
We associate each node with a feature vector $\mathbf{X}_{t,i} \in \mathbb{R}^{C}$ that encodes characteristics of the region, such as appearance (e.g., color).
% 
% The complete node feature tensor is denoted $\mathbf{X} \in \mathbb{R}^{C \times |\mathcal{T}| \times |\mathcal{N}|}$.
% 
Each edge $e_{t,i,j} \in \mathcal{E}$ defines a pairwise spatial connection between the nodes $n_{t,i}$ and $n_{t,j}$.
We associate each edge with a feature vector $\mathbf{A}_{t,i,j} \in \mathbb{R}^{E}$ that encodes pairwise relationships between the associated nodes.
% 
% The complete edge feature tensor is denoted as $\mathbf{A} \in \mathbb{R}^{|\mathcal{T}| \times |\mathcal{N}| \times |\mathcal{N}| \times E}$.
% 
Next, we describe the specific definition of the node and edge sets and feature construction (see Fig.~\ref{fig:method-stgraph}).

\noindent\textbf{Node Regions.}
We define each node $n_{t,i} \in \mathcal{N}$ to represent a fixed region of the facial surface tracked over time.
We represent the facial surface as a sequence of 3D triangular meshes with a fixed topology over time, denoted by
\begin{equation}
    \mathcal{M}_{3D} = (\mathcal{V}, \mathcal{F}).
    \label{eq:mesh_3d}
\end{equation}
Here, $\mathcal{V} \in \mathbb{R}^{|\mathcal{T}| \times V \times 3}$ denotes the set of vertex positions over time and
$\mathcal{F} \in \mathbb{Z}^{F \times 3}$ specifies the fixed triangulation scheme shared across frames.
Here, $\mathcal{V} \in \mathbb{R}^{|\mathcal{T}| \times V \times 3}$ denotes the set of vertex positions over time and $\mathcal{F} \in \mathbb{Z}^{F \times 3}$ specifies the fixed triangulation scheme shared across frames.
For each triangle index $i$, let $f_i \in \mathcal{F}$ be the corresponding triplet of vertex indices, and let $f_{t,i}$ denote its geometric realization at time $t$ under $\mathcal{V}$.
We associate each node $n_{t,i}$ in the graph with $f_{t,i}$, so each node represents a triangular region on the facial surface tracked over time.

\noindent\textbf{Node Features.}
We define the features $\mathbf{X}_{t,i}$ associated with each node $n_{t,i} \in \mathcal{N}$ as the average observable color of its corresponding surface region $f_{t,i}$. 
To ensure that only visible regions contribute, we ignore nodes whose associated face is back-facing with respect to the camera.
Specifically, a node is considered occluded and assigned $\mathbf{X}_{t,i} = \mathbf{0}$ if its face normal $\mathbf{d}_{t,i}$ satisfies $\mathbf{d}_{t,i} \cdot \mathbf{d}_{\text{c}} < 0$, where $\mathbf{d}_{\text{c}} = [0, 0, -1]$ denotes the camera's viewing direction.
For the remaining nodes, let $\mathcal{P}_{t,i}$ be the set of image pixels covered by the projection of $f_{t,i}$ in image space.
The node color is then defined as the mean RGB value across the pixels in $\mathcal{P}_{t,i}$.
This ensures that node features capture surface appearance while remaining stable under motion.

\noindent\textbf{Edge Connectivity}.
To enable localized spatial processing, we define the edge set $\mathcal{E}$ to encode spatial connectivity.
Within each frame $t$, we add an edge between $n_{t,i}$ and $n_{t,j}$ if their corresponding mesh faces share a vertex or if they are the same node, i.e.,
\begin{equation}
    e_{t,i,j} \in \mathcal{E}
    \;\Longleftrightarrow\;
    \bigl(f_{t,i} \cap f_{t,j} \neq \emptyset\bigr) \,\lor\, (i = j).
\label{eq:edge_connectivity}
\end{equation}
We do not instantiate temporal edges; temporal consistency is implicitly enforced through the surface-stable node identities.

\noindent\textbf{Edge features}.
We assign unit weight to all edges in $\mathcal{E}$, this results in a time-invariant adjacency matrix due to the fixed topology, resulting in $\mathbf{A} \in \mathbb{R}^{|\mathcal{N}| \times |\mathcal{N}|}$.
Temporal correspondence is implicitly enforced through the identity-preserving indexing of nodes across time, rather than through explicit temporal edges.

%% file: figures/stgraph.tex
\begin{figure*}
    \centering
    \footnotesize
    \begin{tikzpicture}
        % \fill [fill=blue, opacity=0.10] (0,0) rectangle (\textwidth,0.20\textwidth);

        \node[anchor=south west,inner sep=0] at (0,0)
        {\includegraphics[width=1.00\linewidth]{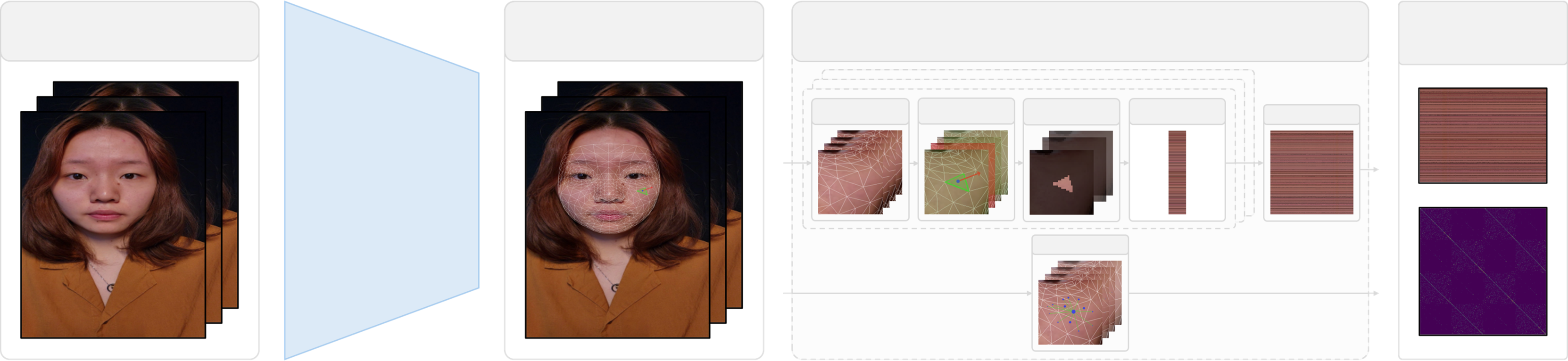}};

        \node[anchor=south west,inner sep=0] at (0.080\linewidth,0.205\linewidth) {$I$};

        \node[anchor=south west,inner sep=0] at (0.210\linewidth,0.090\linewidth) {\shortstack{\itshape \text{3D Facial}\\\itshape \text{Mesh}\\\itshape \text{Detector}}};

        \node[anchor=south west,inner sep=0] at (0.380\linewidth,0.200\linewidth) {$I, \mathcal{M}_{3D}$};

        \node[anchor=south west,inner sep=0] at (0.620\linewidth,0.205\linewidth) {$\mathbf{X}$ and $\mathbf{A}$ Construction};

        \node[anchor=south west,inner sep=0] at (0.922\linewidth,0.200\linewidth) {$(\mathbf{X}, \mathbf{A})$};

        \tiny

        \node[anchor=south west,inner sep=0] at (0.503\linewidth,0.182\linewidth) {$\mathcal{T}$};

        \draw[->] (0.510\linewidth,0.177\linewidth) -- (0.520\linewidth,0.187\linewidth);

        \node[anchor=south west,inner sep=0] at (0.510\linewidth,0.142\linewidth) {$\mathcal{N}$};

        \draw[->] (0.520\linewidth,0.138\linewidth) -- (0.530\linewidth,0.148\linewidth);

        \node[anchor=south west,inner sep=0] at (0.744\linewidth,0.154\linewidth) {$\mathbf{X}_{t}$};

        \node[anchor=south west,inner sep=0] at (0.605\linewidth,0.152\linewidth) {$\mathbf{d}_{t{,}i}$};

        \node[anchor=south west,inner sep=0] at (0.670\linewidth,0.151\linewidth) {$\mathcal{P}_{t{,}i}$};

        \node[anchor=south west,inner sep=0] at (0.536\linewidth,0.152\linewidth) {$f_{t{,}i}$};

        \node[anchor=south west,inner sep=0] at (0.831\linewidth,0.154\linewidth) {$\mathbf{X}$};
        
        \node[anchor=south west,inner sep=0] at (0.672\linewidth,0.069\linewidth) {$e_{t,i,j}$};

        \node[anchor=south west,inner sep=0] at (0.649\linewidth,0.059\linewidth) {$\mathcal{N}$};

        \draw[->] (0.659\linewidth,0.055\linewidth) -- (0.669\linewidth,0.065\linewidth);
        
    \end{tikzpicture}

    \caption{\textbf{STGraph} construction involves obtaining a fixed-topology 3D mesh $\mathcal{M}_{3D}$ for each frame $I_t$.
    Leveraging $\mathcal{M}_{3D}$, we compute for each mesh face $f_{t,i}$ the mean RGB value of its projected pixels and use this as the node feature $\mathbf{X}_{t,i}$ for faces that are front-facing with respect to the camera (faces with $\mathbf{d}_{t,i} \cdot \mathbf{d}_{\text{c}} < 0$ are treated as occluded and assigned $\mathbf{X}_{t,i} = \mathbf{0}$).
    Spatial edges are instantiated between nodes whose corresponding faces share at least one vertex in the canonical UV mesh (plus self-loops), yielding a symmetric, time-invariant adjacency $\mathbf{A}$.
    Stacking the per-face features $\mathbf{X}_{t,i}$ over time together with the fixed adjacency $\mathbf{A}$ defines the facial spatiotemporal graph $\mathcal{G}_{f} = (\mathcal{N}, \mathcal{E}, \mathbf{X}, \mathbf{A})$.}
    \label{fig:method-stgraph}
\end{figure*}
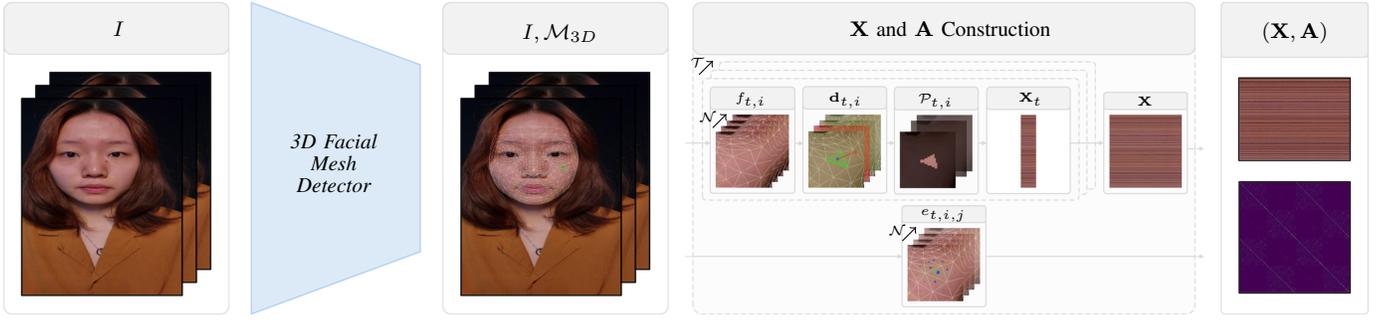

%% file: sec/3b_meshphys.tex
\input{figures/meshphys}

\subsection{MeshPhys}
\label{sec:method-MeshPhys}

We introduce MeshPhys (Fig.~\ref{fig:method-MeshPhys}), a network for estimating waveforms from the STGraph (Sec.~\ref{sec:method-stgraph}).
MeshPhys comprises a spatiotemporal backbone and a waveform prediction head.
The backbone interleaves multi-kernel temporal convolutions, which model multi-scale temporal patterns in node features, with spatial convolutions over the mesh topology, while spatial pooling coarsens the graph by aggregating local features.
The prediction head aggregates the spatial features and projects them onto a waveform estimate $\hat{Y} \in \mathbb{R}^{|\mathcal{T}|}$.

\noindent\textbf{Multi-kernel Temporal Convolution Block}.
The underlying signal has relatively stable temporal structure, but its video appearance varies across datasets due to noise, motion, and recording conditions.
To handle this variability, we use a multi-kernel temporal convolution block that mixes multiple temporal scales in each node's features, with shorter kernels emphasizing high-frequency content and longer kernels providing stronger smoothing under noise.
After input channel projection, we apply $P$ temporal branches in parallel, each consisting of a pointwise convolution followed by a depthwise separable temporal convolution with kernel size $k_{p}$, defined as
\begin{equation}
    \mathbf{Y}_p = \text{TConvModule}_{p}(\mathbf{X}), \quad p = 1,\dots,P.
\label{eq:tconv_branch}
\end{equation}
Here, $\mathbf{X}$ and $\mathbf{Y}_p$ denote the branch input and output, respectively, and $\text{TConvModule}_{p}(\cdot)$ denotes the branch-specific temporal filtering and nonlinearities.
To adaptively weight these temporal scales per clip, we use a clip-wise gating mechanism.
For each branch, we compute a scalar weight from globally pooled features as
\begin{equation}
    \alpha_p = \text{Conv}_{1 \times 1} \!\left( \text{GAP}\! \left( \mathbf{Y}_p \right) \right), 
    \quad p = 1,\dots,P.
    \label{eq:tconv_alpha}
\end{equation}
We then use these weights to gate the branch outputs:
\begin{equation}
    \tilde{\mathbf{Y}}_p
    = \frac{\exp(\alpha_p)}{\sum_{q=1}^{P} \exp(\alpha_q)} \, \mathbf{Y}_p,
    \quad p = 1,\dots,P.
    \label{eq:tconv_gate}
\end{equation}
The weighted branches are then aggregated $\tilde{\mathbf{Y}} = [ \tilde{\mathbf{Y}}_1 \, \| \, \dots \, \| \, \tilde{\mathbf{Y}}_P ]$ and fused through a pointwise convolution, a residual connection from the input is also applied.
%
% \begin{equation}
%     \mathbf{Y} =
%     \phi \! \left( \text{BN} \! \left( \text{Conv}_{1 \times 1} \! \left( \tilde{\mathbf{Y}} \right) \right) \right) + \text{Conv}_{1 \times 1}(\mathbf{X}).
% \label{eq:tconv_output}
% \end{equation}
%
This block therefore, learns a sample-adaptive mixture of temporal scales while remaining parameter-efficient.

\noindent\textbf{Spatial Graph Convolution Block}.
To incorporate spatial modeling, we apply spatial graph convolution~\cite{kipf2016semi} leveraging the spatial adjacency $\mathbf{A}$.
At each time step, node features are linearly projected, aggregated with adjacency-weighted averaging, normalized, and combined with a residual connection.
This operation respects the local topology of the face and enables structured feature propagation across spatially connected regions.

\noindent\textbf{Spatial Graph Pooling Block}.
We coarsen the spatial graph $(\mathbf{X}, \mathbf{A})$ using node clusters~\cite{zhitao2018hierarchical} computed once on canonical vertex positions.
Node features within each cluster are averaged, and clusters are connected if any of their constituent nodes were neighbors in the original mesh.
The use of canonical vertex positions enforces fixed clustering across frames, ensuring temporal aggregation and resultant topology.

%% file: figures/meshphys.tex
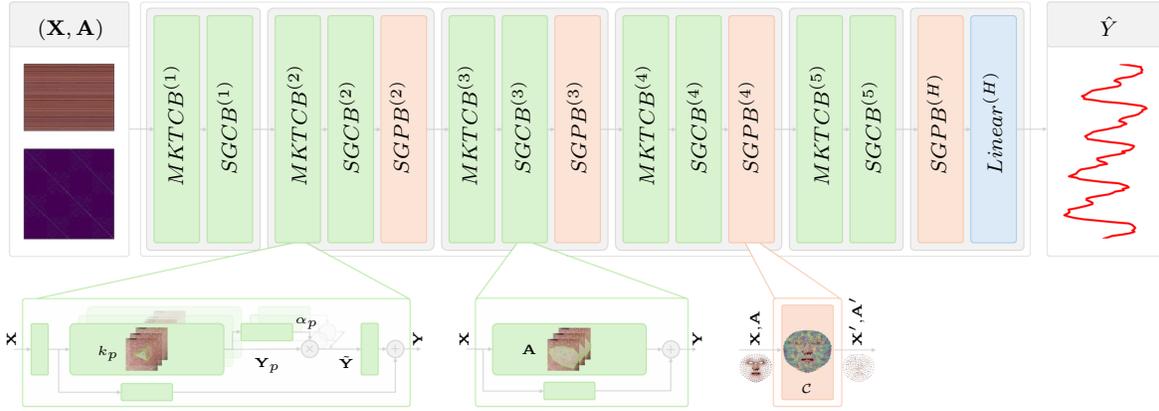
\begin{figure*}
    \centering
    \footnotesize
    \begin{tikzpicture}
        \fill [fill=blue, opacity=0.00] (0,0) rectangle (\textwidth,0.30\textwidth);

        \node at (0.5\textwidth,0.15\textwidth)
        {\includegraphics[width=0.85\textwidth]{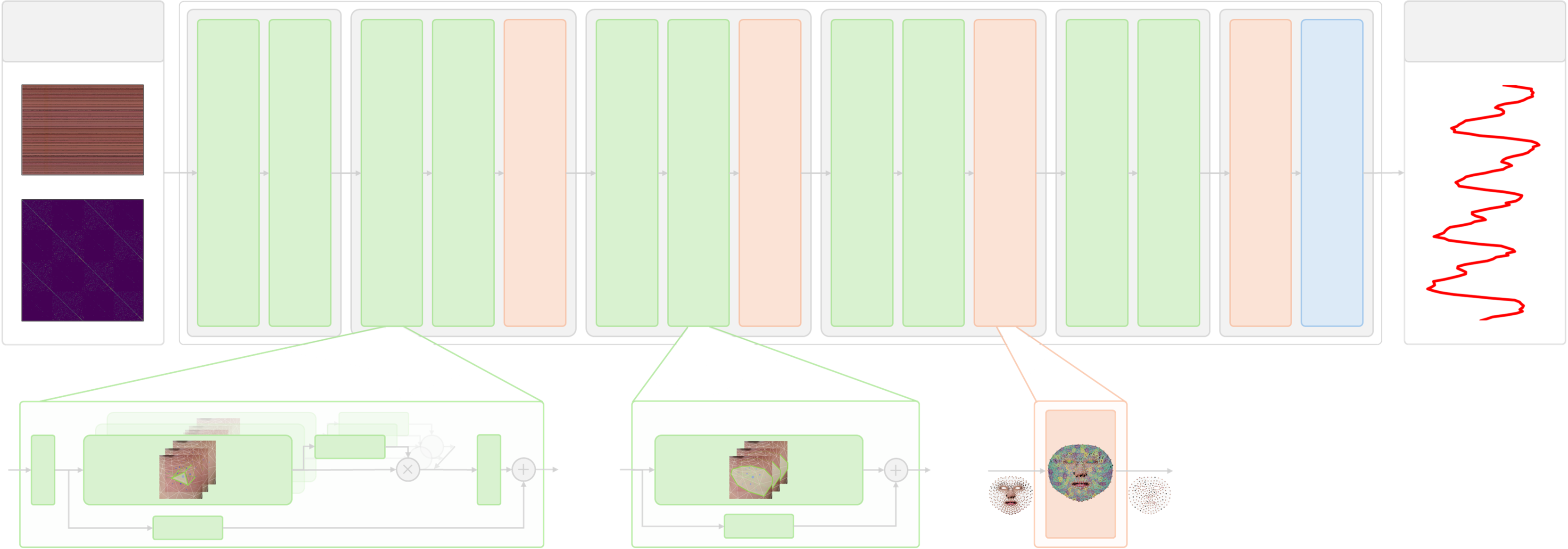}};

        \node[anchor=south west,inner sep=0] (frame) at (0.096\linewidth,0.27\linewidth) {$(\mathbf{X},\mathbf{A})$};

        % Layer 1
        \node[anchor=south west,inner sep=0, rotate=90] (frame) at (0.2035\linewidth,0.163\linewidth) {$MKTCB^{(1)}$};
        \node[anchor=south west,inner sep=0, rotate=90] (frame) at (0.2427\linewidth,0.172\linewidth) {$SGCB^{(1)}$};

        \node[anchor=south west,inner sep=0, rotate=90] (frame) at (0.294\linewidth,0.163\linewidth) {$MKTCB^{(2)}$};
        \node[anchor=south west,inner sep=0, rotate=90] (frame) at (0.3322\linewidth,0.172\linewidth) {$SGCB^{(2)}$};
        \node[anchor=south west,inner sep=0, rotate=90] (frame) at (0.369\linewidth,0.172\linewidth) {$SGPB^{(2)}$};

        \node[anchor=south west,inner sep=0, rotate=90] (frame) at (0.419\linewidth,0.163\linewidth) {$MKTCB^{(3)}$};
        \node[anchor=south west,inner sep=0, rotate=90] (frame) at (0.458\linewidth,0.172\linewidth) {$SGCB^{(3)}$};
        \node[anchor=south west,inner sep=0, rotate=90] (frame) at (0.499\linewidth,0.172\linewidth) {$SGPB^{(3)}$};

        \node[anchor=south west,inner sep=0, rotate=90] (frame) at (0.549\linewidth,0.163\linewidth) {$MKTCB^{(4)}$};
        \node[anchor=south west,inner sep=0, rotate=90] (frame) at (0.587\linewidth,0.172\linewidth) {$SGCB^{(4)}$};
        \node[anchor=south west,inner sep=0, rotate=90] (frame) at (0.625\linewidth,0.172\linewidth) {$SGPB^{(4)}$};

        \node[anchor=south west,inner sep=0, rotate=90] (frame) at (0.677\linewidth,0.163\linewidth) {$MKTCB^{(5)}$};
        \node[anchor=south west,inner sep=0, rotate=90] (frame) at (0.715\linewidth,0.172\linewidth) {$SGCB^{(5)}$};

        \node[anchor=south west,inner sep=0, rotate=90] (frame) at (0.763\linewidth,0.172\linewidth) {$SGPB^{(H)}$};
        \node[anchor=south west,inner sep=0, rotate=90] (frame) at (0.804\linewidth,0.172\linewidth) {$Linear^{(H)}$};

        % PPG
        \node[anchor=south west,inner sep=0] (frame) at (0.875\linewidth,0.273\linewidth) {$\hat{Y}$};

        % MKTCB
        \tiny
        \node[anchor=south west,inner sep=0, rotate=90] (frame) at (0.081\linewidth,0.0455\linewidth) {$\mathbf{X}$};
        \node[anchor=south west,inner sep=0, rotate=0] (frame) at (0.255\linewidth,0.0265\linewidth) {$\mathbf{Y}_{p}$};
        \node[anchor=south west,inner sep=0, rotate=0] (frame) at (0.140\linewidth,0.035\linewidth) {$k_{p}$};
        \node[anchor=south west,inner sep=0, rotate=0] (frame) at (0.285\linewidth,0.0585\linewidth) {$\alpha_{p}$};
        \node[anchor=south west,inner sep=0, rotate=0] (frame) at (0.317\linewidth,0.0295\linewidth) {$\tilde{\mathbf{Y}}$};
        \node[anchor=south west,inner sep=0, rotate=90] (frame) at (0.379\linewidth,0.0455\linewidth) {$\mathbf{Y}$};

        % SGCB
        \node[anchor=south west,inner sep=0, rotate=90] (frame) at (0.413\linewidth,0.0455\linewidth) {$\mathbf{X}$};
        \node[anchor=south west,inner sep=0, rotate=0] (frame) at (0.451\linewidth,0.040\linewidth) {$\mathbf{A}$};
        \node[anchor=south west,inner sep=0, rotate=90] (frame) at (0.584\linewidth,0.0455\linewidth) {$\mathbf{Y}$};

        % SGPB
        \node[anchor=south west,inner sep=0, rotate=90] (frame) at (0.628\linewidth,0.0455\linewidth) {$\mathbf{X}{,}\mathbf{A}$};
        \node[anchor=south west,inner sep=0, rotate=0] (frame) at (0.656\linewidth,0.011\linewidth) {$\mathcal{C}$};
        \node[anchor=south west,inner sep=0, rotate=90] (frame) at (0.703\linewidth,0.0455\linewidth) {$\mathbf{X}'{,}\mathbf{A}'$};

    \end{tikzpicture}
    \caption{
    \textbf{MeshPhys} operates on the \textbf{STGraph} using a spatiotemporal graph-convolutional network.
    The MeshPhys backbone extracts and models localized spatiotemporal features encoded on the facial surface through the STGraph and project these features to a waveform estimate through a simple linear prediction head.
    The backbone consists of five-layers, with each layer comprised of a multi-kernel temporal node-wise convolution block ($MKTCB$), a spatial graph convolution block ($SGCB$), and optionally a spatial graph pooling block ($SGPB$).
    MeshPhys uses 16 base channels and expands the channel dimension to 32, 64, 128, and 128 in subsequent layers.
    Spatial pooling is specifically applied in layers 2, 3, and 4, with global spatial pooling performed in the prediction head.
    Combined, this effectively constrains spatiotemporal modeling to the facial surface.}
    \label{fig:method-MeshPhys}
\end{figure*}

%% file: sec/3c_losses.tex
\subsection{Training Objective}
\label{sec:method-lossfunction}
We supervise rPPG waveform estimation with a composite objective that combines morphological correlation, shift-invariant alignment, SNR-based sample weighting, and PR-band spatial coherence, given by
\begin{equation}
    \mathcal{L} = w_{\text{SNR}} \cdot (\lambda_{0} \mathcal{L}_{\rho}^{\varphi} + \lambda_{1} \mathcal{L}_{\Delta \rho}^{\varphi} + \lambda_{2} \mathcal{L}_{\Delta^{2} \rho}^{\varphi}) + \lambda_{3} \mathcal{L}_{\text{GS}}^{\mathcal{B}}.
    \label{eq:training_objective}
\end{equation}
Next, we detail the individual components.

\noindent\textbf{Morphological Correlation Losses}.
We adopt correlation-based objectives, following~\cite{yu2019physnet}, to capture scale- and offset-invariant similarity between the predicted waveform $\hat{Y}$ and reference waveform $Y$.
To better capture morphology, we also correlate first- and second-order temporal derivatives.
Let $\Delta Y_t = Y_t - Y_{t-1}$ and $\Delta^2 Y_t = Y_{t+1} - 2Y_t + Y_{t-1}$, and denote by $\rho(\cdot,\cdot)$ the Pearson correlation.
The morphological losses are then
\begin{equation}
\begin{aligned}
    \mathcal{L}_{\rho} &= 1 - \rho(\hat{Y}, Y), \\
    \mathcal{L}_{\Delta \rho} &= 1 - \rho(\Delta \hat{Y}, \Delta Y), \\
    \mathcal{L}_{\Delta^{2} \rho} &= 1 - \rho(\Delta^2 \hat{Y}, \Delta^2 Y).
\end{aligned}
\label{eq:morph_losses}
\end{equation}

\noindent\textbf{Phase-Shift Alignment}.
Latency and synchronization drift induce sample-wise phase shifts between video-based BVP and reference PPG, making strictly aligned supervision unreliable~\cite{zhan2020sensitivity,comas2022talos}.
We therefore compute an phase-shift robust version of any base loss $\mathcal{L}(\hat{Y},Y)$ by averaging over circular shifts~\cite{gideon2021waytomyheart} as
\begin{equation}
    \mathcal{L}^{\varphi}(\hat{Y}, Y) =
    \sum_{\varphi = -\Phi}^{\Phi} w(\varphi)\, \mathcal{L} ( \hat{Y}_{\varphi}, Y )
    \label{eq:phase_robust_loss}
\end{equation}
where $\hat{Y}_{\varphi}$ is $\hat{Y}$ circularly shifted by $\varphi$ frames and $\Phi$ is the maximum offset.
The weights in \eqref{eq:phase_robust_loss} are given by
\begin{equation}
    w(\varphi) =
    \frac{\exp \left( -\tau \cdot \mathcal{L} ( \hat{Y}_{\varphi}, Y ) \right)}
         {\sum_{\varphi'} \exp \left( -\tau \cdot \mathcal{L} ( \hat{Y}_{\varphi'}, Y ) \right)}
    \label{eq:phase_weights}
\end{equation}
where $\tau$ controls the soft-min sharpness.
We apply this transformation to each morphological loss to define $\mathcal{L}_{\rho}^{\varphi}$, $\mathcal{L}_{\Delta\rho}^{\varphi}$, and $\mathcal{L}_{\Delta^{2}\rho}^{\varphi}$, with $\Phi=50$ and $\tau=10.0$.

\noindent\textbf{Signal-to-Noise Ratio (SNR) Weighting}.
To down-weight samples with unreliable supervision, we compute a weight from the SNR of the reference PPG.
We estimate the power spectral density $P(f)$ of $Y$ (Welch’s method) and measure how concentrated it is
around the reference pulse rate $f_{\text{PR}}$ using
\begin{equation}
    G(f) = \exp\left(-\frac{(f - f_{\text{PR}})^2}{2\sigma^2}\right),
\label{eq:snr_gaussian}
\end{equation}
and define the SNR-based sample weight as
\begin{equation}
    w_{\text{SNR}} = \left( \frac{\int G(f) P(f)\, df}{\int_{\mathcal{B}} P(f)\, df} \right)^{\gamma},
\label{eq:snr_weight}
\end{equation}
where $\mathcal{B} = [f_{\text{min}}, f_{\text{max}}]$ is the physiological band (0.5–3.0 Hz), $\sigma$ controls the window width, and $\gamma = 0.50$.
We clip $w_{\text{SNR}}$ below at $0.10$.

\noindent\textbf{Graph Smoothness Loss}.
To encourage spatially coherent PR-band activity on the facial graph, we apply a graph Laplacian regularizer to late-layer node features band-pass filtered in the physiological range, given by
\begin{equation}
    \mathcal{L}_{\text{GS}}^{\mathcal{B}}
    = \frac{1}{C T V} \sum_{i,t} \tilde{z}_{i,t}^{(l)\top} L^{(l)} \tilde{z}_{i,t}^{(l)}.
\label{eq:graph_smoothness}
\end{equation}
Here, $\tilde{z}_{i,t}^{(l)} \in \mathbb{R}^{|\mathcal{N}^{(l)}|}$ is the feature vector of node $i$
at frame $t$ in layer $l$ after band-pass filtering to $\mathcal{B}$, $L^{(l)}$ is the
unnormalized Laplacian constructed from $\mathbf{A}^{(l)}$, and $C$, $T$, and $V$ denote
the number of channels, frames, and nodes, respectively.
We apply this regularizer at the final backbone layer ($l=4$).

%% file: sec/4_experiments.tex
\section{Experiments}
\label{sec:exp}
To assess the performance and generalization of this method, we conduct comprehensive experiments for rPPG-based pulse rate (PR) estimation on four publicly available datasets: PURE~\cite{stricker2014pure}, UBFC-rPPG~\cite{bobbia2019ubfc}, MMPD~\cite{tang2023mmpd}, and VIPL-HR~\cite{niu2019vipl}\footnote{The relevant Institutional Review Boards approved experiments involving human subjects described in this paper.}. 
PURE and UBFC-rPPG are small-scale datasets, containing 59 and 42 videos, respectively, collected under constrained conditions.
The MMPD dataset is a medium-scale, complex challenging dataset containing a variety of motion and lighting scenarios across 660 videos.
The VIPL-HR dataset is a large-scale dataset containing 2,378 videos captured across a variety of subjects and video recording devices. 
These datasets span a wide range of capture conditions, motion complexity, and subject diversity, making them suitable for benchmarking both model capacity and generalization.

\input{sec/4a_setup}
\input{sec/4b_intra}
\input{sec/4c_cross}
\input{sec/4d_ablation_stgraph}
\input{sec/4e_ablation_meshphys}

%% file: sec/4a_setup.tex
\subsection{Experimental Setup}
\label{sec:experiments-setup}

MeshPhys employs $P{=}3$ branches in the multi-kernel temporal block with kernel sizes $k \in \{3,5,9\}$ and each spatial graph pooling stage down-samples the number of nodes by a factor of $4$.
We use the AdamW optimizer (learning rate $5\times 10^{-4}$, batch size $16$) and train for $100$ epochs, using the composite objective in Sec.~\ref{sec:method-lossfunction} with weights $\lambda_{0}=1.00$, $\lambda_{1}=0.50$, $\lambda_{2}=0.10$, and $\lambda_{3}=0.002$.
For well-synchronized datasets (PURE, UBFC, MMPD) we disable phase-shift alignment by setting $\Phi=0$.
For each experiment, we select the checkpoint with the lowest validation loss for testing.
We evaluate models on pulse-rate (PR) estimation following the protocols described in the following sections.
For each video, we aggregate the predicted waveform $\hat{Y}$ over all clips, apply a 3rd-order Butterworth band-pass filter with cut-off frequencies $\mathcal{B}$, compute the Welch power spectral density $\hat{P}(f)$, and take the dominant frequency in $\mathcal{B}$ as the estimated PR, $\hat{f}_{\text{PR}}$.
We apply the same procedure to the reference waveform $Y$ to obtain $f_{\text{PR}}$.
We report mean absolute error (MAE) and root mean square error (RMSE) in beats-per-minute (BPM), together with the Pearson correlation coefficient between $f_{\text{PR}}$ and $\hat{f}_{\text{PR}}$, averaging all metrics over three seeds.
Further details on data processing, STGraph construction, and data augmentation prior to training are provided in the supplement.

%% file: sec/4b_intra.tex
\input{figures/ba_mmpd}

\input{tables/intra}

\subsection{Intra-dataset Evaluation}
\label{sec:experiments-intra}
Table~\ref{tab:experiments-intra} demonstrates that MeshPhys achieves strong intra-dataset performance, particularly on high-variance datasets such as MMPD and VIPL-HR, which contain greater motion, illumination, and physiological variability.
Compared to video-based methods such as RhythmFormer, MeshPhys achieves improved performance on MMPD (MAE reduction of 9.8\%) and VIPL-HR (MAE reduction of 21.7\%) while using 85.9\% fewer parameters, demonstrating the effectiveness of a surface-aligned inductive bias over conventional video-based representations.
On VIPL-HR, MeshPhys outperforms STMap-based methods NEST (MAE reduction of 25.8\%) and rPPG-MAE (MAE reduction of 21.9\%) while requiring 96.6\% and 99.5\% fewer parameters, respectively, and performs competitively with RhythmGaussian.
On constrained datasets (PURE, UBFC-rPPG), MeshPhys also achieves state-of-the-art or near state-of-the-art scores, confirming its strong performance across both simple and complex data regimes.
These results suggest that the surface-aligned STGraph provides a compact yet expressive representation, outperforming both conventional video pipelines and high-capacity STMap models in terms of robustness and efficiency.
Although lightweight, MeshPhys remains prone to overfitting without strong regularization, indicating that the STGraph makes surface-relevant features more directly accessible and easier to memorize.
Together, these findings support our central claim that constraining the model’s spatiotemporal processing to the 3D facial surface constitutes a strong prior for facial rPPG estimation.
Evaluation protocols follow prior work on each dataset~\cite{spetlik2018visual,song2021pulsegan,zou2024rhythmformer,niu2019rhythmnet}.

\input{figures/signal_mmpd}

MeshPhys achieves strong agreement with ground-truth pulse rates, as evidenced by tight clustering along the identity line and narrow Bland--Altman limits in Fig.~\ref{fig:results-baa}. 
The low mean bias and evenly distributed residuals across a broad BPM range indicate stable performance without systematic drift or collapse at the extremes.
Notably, outlier predictions are isolated rather than clustered, suggesting resilience to variation in pulse rate magnitude and signal characteristics.
A representative example in Fig.~\ref{fig:results-timeseries} shows that the predicted waveform retains key temporal features and closely matches the ground truth in both time and frequency domains, with only minor under-attenuation of high-frequency detail.
The preservation of dominant spectral energy confirms that the model reliably captures the core periodicity critical for pulse-rate inference.
These results support the reliability and physiological fidelity of MeshPhys under intra-dataset evaluation.

%% file: figures/ba_mmpd.tex
\input{data/ba_stats.tex}

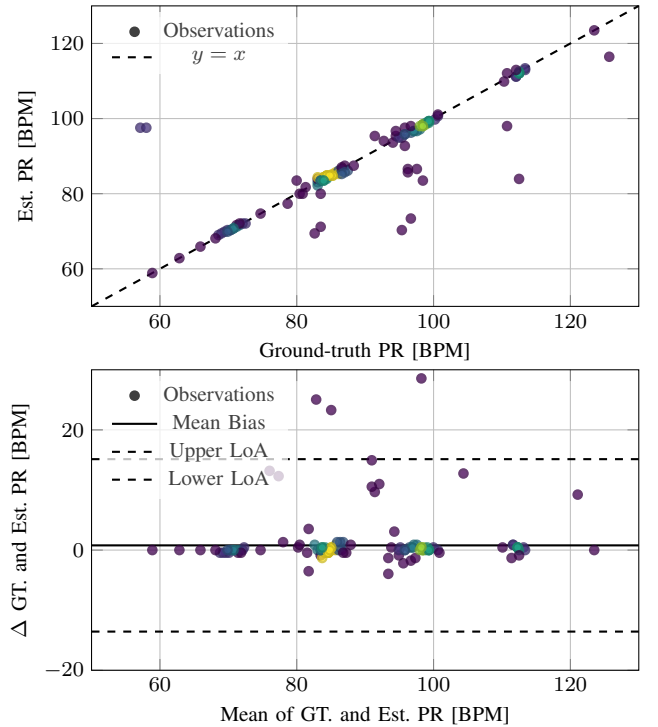
\begin{figure}
    \centering
    \footnotesize
    \begin{tikzpicture}
        \begin{groupplot}[
            group style={
                group size=1 by 2,      % 1 column, 2 rows
                vertical sep=24pt
            },
            width=\linewidth, 
            height=0.63\linewidth,    % per-axis size (match your originals)
            grid=both,
            every axis plot/.append style={line width=0.8pt},
            legend style={
                at={(rel axis cs:0.02,0.98)},anchor=north west,
                draw=none, fill=white, fill opacity=0.75
            },
            legend image post style={opacity=1}
        ]
        
        % ===================== Row 1: Scatter (y=x) =====================
        \nextgroupplot[
            xlabel={Ground-truth PR [BPM]},
            ylabel={Est. PR [BPM]},
            xmin=50, 
            xmax=130, 
            ymin=50, 
            ymax=130
        ]
        % points (density coloring optional)
        \addplot[
            only marks, 
            mark=*,
            mark size=1.6pt,
            scatter, 
            scatter src=explicit,
            visualization depends on=\thisrow{density} \as \dens,
            colormap/viridis,
            opacity=0.75,
            mark options={
                fill opacity=0.75, 
                draw opacity=0.75
            }
        ] table[
            col sep=comma,
            x=gold, 
            y=new,
            meta=density % comment out if no 'density' column
        ] {data/ba_data.csv};
        \addlegendentry{Observations}
        
        % identity line
        \addplot[
            black, dashed,
            domain=\pgfkeysvalueof{/pgfplots/xmin}:\pgfkeysvalueof{/pgfplots/xmax}
        ] {x};
        \addlegendentry{$y=x$}
        
        % ===================== Row 2: Bland–Altman =====================
        \nextgroupplot[
            xlabel={Mean of GT. and Est. PR [BPM]},
            ylabel={$\Delta$ GT. and Est. PR [BPM]},
            xmin=50, 
            xmax=130, 
            ymin=-20, 
            ymax=30
        ]
        % points (density coloring optional)
        \addplot[
            only marks, mark=*,
            mark size=1.6pt,
            scatter, scatter src=explicit,
            visualization depends on=\thisrow{density} \as \dens,
            colormap/viridis,
            opacity=0.75,
            mark options={fill opacity=0.75, draw opacity=0.75}
        ] table[
            col sep=comma,
            x=mean, y=diff,
            meta=density % comment out if no 'density' column
        ] {data/ba_data.csv};
        \addlegendentry{Observations}
        
        % BA lines
        \addplot[black, thick,
            domain=\pgfkeysvalueof{/pgfplots/xmin}:\pgfkeysvalueof{/pgfplots/xmax}
        ] {\BAmean};
        \addlegendentry{Mean Bias}
        
        \addplot[black, dashed,
            domain=\pgfkeysvalueof{/pgfplots/xmin}:\pgfkeysvalueof{/pgfplots/xmax}
        ] {\BAloaUpper};
        \addlegendentry{Upper LoA}
        
        \addplot[black, dashed,
            domain=\pgfkeysvalueof{/pgfplots/xmin}:\pgfkeysvalueof{/pgfplots/xmax}
        ] {\BAloaLower};
        \addlegendentry{Lower LoA}
        
        \end{groupplot}
    \end{tikzpicture}
    
    \caption{Comparison between the estimated and ground-truth pulse rates from intra-dataset evaluation on MMPD.
    Scatter plot (top) of the estimated versus ground-truth PR, with the identity line indicating perfect agreement. 
    Bland–Altman plot (bottom) showing the difference between the estimated and ground-truth PR as a function of their mean, along with the mean bias and 95\% limits of agreement.}
    \label{fig:results-baa}
\end{figure}

%% file: data/ba_stats.tex
% Auto-generated BA stats
\pgfmathsetmacro{\BAmean}{0.7828913}
\pgfmathsetmacro{\BAloaUpper}{15.1284}
\pgfmathsetmacro{\BAloaLower}{-13.56262}
\pgfmathsetmacro{\BAcorr}{0.8601418}
\def\BAunits{bpm}

%% file: tables/intra.tex
\begin{table*}
    \centering
    \caption{Intra-dataset PR estimation performance of models on the PURE, UBFC-rPPG, MMPD, and VIPL-HR datasets. 
    Input representations are denoted as: $\blacktriangle$ ROI, $\blacklozenge$ STMaps, $\blacksquare$ video, and $\bigstar$ STGraph respectively. 
    We denote traditional methods with $^{\circ}$, self-supervised methods with $^{*}$, and leave supervised methods un-denoted.
    The best and second best results are formatted as \textbf{bold} and \underline{underline} respectively.}
    \footnotesize
    \begin{tabular}{l c c c c c c c c c c c c c}
        \toprule

        & \multicolumn{3}{c}{\textbf{PURE}} & \multicolumn{3}{c}{\textbf{UBFC-rPPG}} & \multicolumn{3}{c}{\textbf{MMPD}} & \multicolumn{3}{c}{\textbf{VIPL-HR}} \\

        \cline{2-4}
        \cline{5-7}
        \cline{8-10}
        \cline{11-13}
        
        \multirow{2}{*}{Method} & MAE $\downarrow$ & RMSE $\downarrow$ & \multirow{2}{*}{\textit{r} $\uparrow$} & MAE $\downarrow$ & RMSE $\downarrow$ & \multirow{2}{*}{\textit{r} $\uparrow$} & MAE $\downarrow$ & RMSE $\downarrow$ & \multirow{2}{*}{\textit{r} $\uparrow$} & MAE $\downarrow$ & RMSE $\downarrow$ & \multirow{2}{*}{\textit{r} $\uparrow$} \\
        
        & (BPM) & (BPM) & & (BPM) & (BPM) & & (BPM) & (BPM) & & (BPM) & (BPM) & \\

        \midrule

        CHROM~\cite{de2013robust} $\blacktriangle^{\circ}$ & 5.77 & 14.93 & 0.81 & 4.06 & 8.83 & 0.89 & 13.66 & 18.76 & 0.08 & 11.44 & 16.97 & 0.28 \\
        
        POS~\cite{wang2016algorithmic} $\blacktriangle^{\circ}$ & 3.67 & 11.82 & 0.88 & 4.08 & 7.72 & 0.92 & 12.36 & 17.71 & 0.18 & 11.5 & 17.2 & 0.30 \\

        \midrule

        RhythmNet~\cite{niu2019rhythmnet} $\blacklozenge$ & - & - & - & 1.19 & 2.10 & 0.98 & - & - & - & 5.30 & 8.14 & 0.76 \\
        
        NAS-HR~\cite{lu2021nashr} $\blacklozenge$ & - & - & - & - & - & - & - & - & - & 5.12 & 8.01 & 0.79 \\
        
        CVD~\cite{niu2020video} $\blacklozenge$ & 1.29 & 2.01 & 0.98 & 2.19 & 3.12 & \underline{0.99} & - & - & - & 5.02 & 7.97 & 0.79 \\
        
        Dual-GAN~\cite{lu2021dual} $\blacklozenge$ & 0.82 & 1.31 & \underline{0.99} & 0.44 & 0.67 & \underline{0.99} & - & - & - & 4.93 & 7.68 & 0.81 \\
        
        BVPNet~\cite{das2021bvpnet} $\blacklozenge$ & - & - & - & - & - & - & - & - & - & 5.34 & 7.85 & 0.70 \\
        
        Dual-TL~\cite{qian2024dual} $\blacklozenge$ & 0.37 & 0.68 & 0.99 & \underline{0.17} & 0.41 & 0.99 & - & - & - & 4.36 & \underline{6.92} & 0.69 \\
        
        rPPG-MAE~\cite{liu2024rppg} $\blacklozenge^{*}$ & 0.40 & 0.92 & \underline{0.99} & \underline{0.17} & \textbf{0.21} & \underline{0.99} & - & - & - & 4.52 & 7.49 & 0.81 \\
        
        NEST~\cite{lu2023nest} $\blacklozenge$ & - & - & - & - & - & - & - & - & - & 4.76 & 7.51 & \underline{0.84} \\
        
        RhythmGaussian~\cite{lu2025rhythmgaussian} $\blacklozenge^{*}$ & - & - & - & - & - & - & - & - & - & \underline{4.22} & 7.12 & \textbf{0.92} \\

        \midrule

        DeepPhys~\cite{chen2018deepphys} $\blacksquare$ & 0.83 & 1.54 & \underline{0.99} & 6.27 & 10.82 & 0.65 & 22.27 & 28.92 & -0.03 & 11.0 & 13.8 & 0.11 \\
        
        PhysNet~\cite{yu2019physnet} $\blacksquare$ & 2.10 & 2.60 & \underline{0.99} & 2.95 & 3.67 & 0.97 & 4.80 & 11.80 & 0.60 & 10.8 & 14.8 & 0.20 \\
        
        TS-CAN~\cite{liu2020multi} $\blacksquare$ & 2.48 & 9.01 & 0.92 & 1.70 & 2.72 & \underline{0.99} & 9.71 & 17.22 & 0.44 & - & - & - \\
        
        PhysFormer~\cite{yu2022physformer} $\blacksquare$ & 1.10 & 1.75 & \underline{0.99} & 0.50 & 0.71 & \underline{0.99} & 11.99 & 18.41 & 0.18 & 4.97 & 7.79 & 0.78 \\
        
        PhysFormer++~\cite{yu2023physformerpp} $\blacksquare$ & - & - & - & - & - & - & - & - & - & 4.88 & 7.62 & 0.80 \\
        
        EfficientPhys~\cite{liu2023efficientphys} $\blacksquare$ & - & - & - & 1.14 & 1.81 & \underline{0.99} & 13.47 & 21.32 & 0.21 & - & - & - \\
        
        PhysGCN~\cite{shao2023hyperbolic} $\blacksquare$ & - & - & - & 4.64 & 7.37 & 0.87 & - & - & - & 4.70 & 7.44 & 0.82 \\
        
        RhythmFormer~\cite{zou2024rhythmformer} $\blacksquare$ & \underline{0.27} & \underline{0.47} & \underline{0.99} & 0.50 & 0.78 & \underline{0.99} & \underline{3.07} & \textbf{6.81} & \textbf{0.86} & 4.51 & 7.98 & 0.78 \\
        
        RhythmMamba~\cite{zou2024rhythmmamba} $\blacksquare$ & \textbf{0.23} & \textbf{0.34} & \underline{0.99} & 0.50 & 0.75 & \underline{0.99} & 3.16 & \underline{7.27} & 0.84 & 4.30 & 7.49 & 0.81 \\
        
        GraphPhys~\cite{xiong2024graphphys} $\blacksquare$ & - & - & - & - & - & - & - & - & - & 6.69 & 9.70 & 0.48 \\
        
        STGNet~\cite{stgnet2024xiong} $\blacksquare$ & 0.69 & 0.98 & \underline{0.99} & 0.45 & 0.58 & \underline{0.99} & - & - & - & - & - & - \\    

        CodePhys~\cite{codephys2025} $\blacksquare$ & 0.39 & 0.83 & 0.99 & 0.21 & 0.26 & 0.99 & - & - & - & 4.27 & 7.11 & 0.81 \\

        \midrule

        \rowcolor{gray!25} MeshPhys $\bigstar$ & \underline{0.27} & 0.48 & \textbf{1.00} & \textbf{0.15} & \underline{0.25} & \textbf{1.00} & \textbf{2.77} & 7.50 & \underline{0.85} & \textbf{3.53} & \textbf{6.82} & \underline{0.84} \\

        \bottomrule
        
    \end{tabular}

    \label{tab:experiments-intra}
\end{table*}

%% file: figures/signal_mmpd.tex
\begin{figure}
    \centering
    \footnotesize
    \begin{tikzpicture}
        \begin{groupplot}[
            group style={
                group size=1 by 2,            % 1 column, 2 rows
                vertical sep=24pt              % gap between axes
            },
            width=\linewidth,
            height=0.45\linewidth,
            grid=both,
            every axis plot/.append style={line width=0.8pt},
            legend style={
                at={(rel axis cs:0.98,0.98)},
                anchor=north east,
                fill=white, 
                fill opacity=0.8, 
                draw=none
            },
            legend image post style={opacity=1},
        ]
        
        % --- Row 1: Time series ---
        \nextgroupplot[
          xlabel={Time (s)}, 
          ylabel={PPG Signal (a.u.)},
          restrict x to domain=0:15, 
          unbounded coords=discard,
          xmin=0, 
          xmax=15,
          ymin=-2.0,
          ymax=2.5
        ]
        \addplot+[no marks, opacity=0.75] 
            table[col sep=comma, x=t_s, y=gt_ppg] {data/bvp_timeseries.csv};
        \addlegendentry{Ground-truth}
        
        \addplot+[no marks, opacity=0.75] 
            table[col sep=comma, x=t_s, y=pred_ppg] {data/bvp_timeseries.csv};
        \addlegendentry{Prediction}
        
        % --- Row 2: Spectrum ---
        \nextgroupplot[
          xlabel={Frequency (BPM)}, 
          ylabel={Power (dB)},
          xmin=30, 
          xmax=240, 
          ymin=-40, 
          ymax=10,
        ]
        \addplot+[no marks, opacity=0.75] 
            table[col sep=comma, x=f_bpm, y=gt_db] {data/bvp_spectrum.csv};
        \addlegendentry{Ground-truth}
        
        \addplot+[no marks, opacity=0.75] 
            table[col sep=comma, x=f_bpm, y=pred_db] {data/bvp_spectrum.csv};
        \addlegendentry{Prediction}
        
        \end{groupplot}
    \end{tikzpicture}
    \caption{Visualization of the predicted and ground-truth PPG signals in the time domain (top) and frequency domain (bottom) for a representative sample from the test set from intra-dataset evaluation on MMPD. 
    The selected sample corresponds to the median absolute pulse rate error on the test set.}
    \label{fig:results-timeseries}
\end{figure}
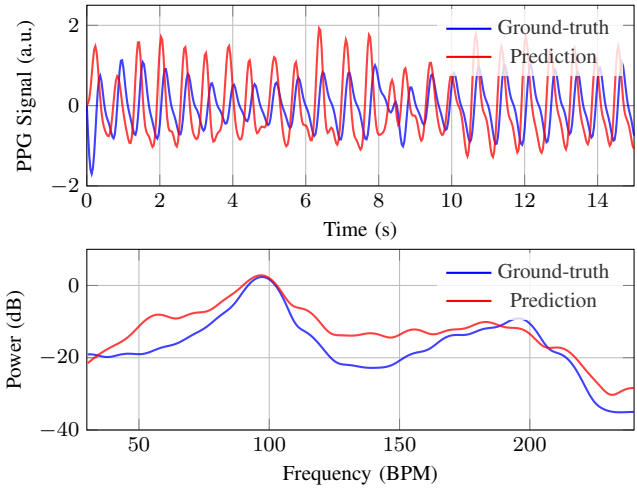

%% file: sec/4c_cross.tex
\input{tables/ssdg}

\subsection{Single-Source Domain Generalization Evaluation}
\label{sec:experiments-cross}
Table~\ref{tab:experiments-cross} presents performance under the single-source domain generalization (SSDG) setting, where models are trained on a single dataset and evaluated on unseen domains.
This setup isolates robustness under domain shift and serves as a strong test of whether a model's inductive bias aligns with the underlying task rather than dataset-specific artifacts.
MeshPhys achieves state-of-the-art performance across all SSDG transfers, including challenging cases such as training on PURE or UBFC-rPPG and testing on MMPD, despite being trained on low-variance data.
It consistently outperforms much larger video-based models such as RhythmFormer in both MAE and Pearson correlation $r$, highlighting robustness to shifts in motion, pose, and illumination.
These results indicate that the STGraph representation robustly encodes features relevant for rPPG estimation.
Compared to STMap-based models (e.g., NEST, rPPG-MAE, Dual-TL), MeshPhys generalizes substantially better—particularly on MMPD, where most STMap baselines are either unreported or degrade markedly under complex domain shift.
This suggests that the STGraph offers a stronger task-aligned inductive bias than STMaps alone, enabling more effective spatiotemporal modeling through explicit alignment with the facial surface, the spatial support of the physiological signal.
These results reinforce our central hypothesis that enforcing surface-aligned modeling at the input level provides a task-aligned inductive bias that supports both performance and generalization across diverse domains.
Evaluation protocols follow prior work~\cite{liu2023rppgtoolboxdeepremoteppg,zou2024rhythmformer}.

%% file: tables/ssdg.tex
\begin{table*}
    \centering
    \caption{Single-source domain generalization (SSDG) PR estimation performance of models trained on PURE/UBFC-rPPG and tested on PURE/UBFC-rPPG/MMPD. 
    Input representations are denoted as: $\blacktriangle$ ROI, $\blacklozenge$ STMaps, $\blacksquare$ video, and $\bigstar$ STGraph respectively. 
    We denote un-supervised methods with $^{\circ}$, self-supervised methods with $^{*}$, and leave supervised methods un-denoted. 
    The best and second best results are formatted as \textbf{bold} and \underline{underline} respectively.}
    \footnotesize
    \begin{tabular}{l c c c c c c c c c c c c c c c c c}
        \toprule

        & \multicolumn{3}{c}{\textbf{PURE} $\rightarrow$ \textbf{UBFC-rPPG}} & \multicolumn{3}{c}{\textbf{PURE} $\rightarrow$ \textbf{MMPD}} & \multicolumn{3}{c}{\textbf{UBFC-rPPG} $\rightarrow$ \textbf{PURE}} & \multicolumn{3}{c}{\textbf{UBFC-rPPG} $\rightarrow$ \textbf{MMPD}}  \\

        \cline{2-4}
        \cline{5-7}
        \cline{8-10}
        \cline{11-13}
        
        \multirow{2}{*}{Method} & MAE $\downarrow$ & RMSE $\downarrow$ & \multirow{2}{*}{\textit{r} $\uparrow$} & MAE $\downarrow$ & RMSE $\downarrow$ & \multirow{2}{*}{\textit{r} $\uparrow$} & MAE $\downarrow$ & RMSE $\downarrow$ & \multirow{2}{*}{\textit{r} $\uparrow$} & MAE $\downarrow$ & RMSE $\downarrow$ & \multirow{2}{*}{\textit{r} $\uparrow$} \\
        & (BPM) & (BPM) & & (BPM) & (BPM) & & (BPM) & (BPM) & & (BPM) & (BPM) & \\

        \midrule

        CHROM~\cite{de2013robust} $\blacktriangle^{\circ}$ & 5.77 & 14.93 & 0.81 & 13.66 & 18.76 & 0.08 & 4.06 & 8.83 & 0.89 & 13.66 & 18.76 & 0.08 \\
        
        POS~\cite{wang2016algorithmic} $\blacktriangle^{\circ}$ & 3.67 & 11.82 & 0.88 & 12.36 & 17.71 & 0.18 & 4.08 & 7.72 & 0.92 & 12.36 & 17.71 & 0.18 \\

        \midrule

        Dual-GAN~\cite{lu2021dual} $\blacklozenge$ & 0.74 & 1.02 & 1.00 & - & - & - & - & - & - & - & - & - \\
        
        Dual-TL~\cite{qian2024dual} $\blacklozenge$ & - & - & - & - & - & - & 14.12 & 23.47 & 0.23 & - & - & - \\
        
        rPPG-MAE~\cite{liu2024rppg} $\blacklozenge^{*}$ & 1.28 & 2.75 & - & - & - & - & 13.55 & 20.27 & - & - & - & - \\
        
        NEST~\cite{lu2023nest} $\blacklozenge$ & - & - & - & - & - & - & 6.07 & 9.06 & 0.76 & - & - & - \\

        \midrule

        DeepPhys~\cite{chen2018deepphys} $\blacksquare$ & 1.21 & 2.90 & \underline{0.99} & 16.92 & 24.61 & 0.05 & 5.54 & 18.51 & 0.66 & 17.50 & 25.00 & 0.05  \\
        
        PhysNet~\cite{yu2019physnet} $\blacksquare$ & 0.98 & 2.48 & \underline{0.99} & 13.22 & 19.61 & 0.23 & 8.06 & 19.71 & 0.61 & 10.24 & 16.54 & 0.29  \\
        
        TS-CAN~\cite{liu2020multi} $\blacksquare$ & 1.30 & 2.87 & \underline{0.99} & 13.94 & 21.61 & 0.20 & 3.69 & 13.80 & 0.82 & 14.01 & 21.04 & 0.24  \\
        
        PhysFormer~\cite{yu2022physformer} $\blacksquare$ & 1.44 & 3.77 & 0.98 & 14.57 & 20.71 & 0.15 & 12.92 & 24.36 & 0.47 & 12.10 & 17.79 & 0.17   \\
        
        EfficientPhys~\cite{liu2023efficientphys} $\blacksquare$ & 2.07 & 6.32 & 0.94 & 14.03 & 21.62 & 0.17 & 5.47 & 17.04 & 0.71 & 13.78 & 22.25 & 0.09   \\
        
        Spiking-Phys~\cite{liu2025spiking} $\blacksquare$ & 2.80 & - & 0.95 & 14.57 & - & 0.14 & 3.83 & - & 0.83 & 14.15 & - & 0.15   \\
        
        RhythmFormer~\cite{zou2024rhythmformer} $\blacksquare$ & \underline{0.89} & \underline{1.83} & \underline{0.99} & 8.98 & \textbf{14.85} & \textbf{0.51} & \underline{0.97} & \underline{3.36} & \underline{0.99} & 9.08 & \underline{15.07} & \underline{0.53}   \\
        
        RhythmMamba~\cite{zou2024rhythmmamba} $\blacksquare$ & 0.95 & \underline{1.83} & \underline{0.99} & 10.44 & 16.70 & 0.36 & 1.98 & 6.51 & 0.96 & 10.63 & 17.14 & 0.34 \\

        BeatFormer~\cite{comas2025beatformer} $\blacksquare$ & - & - & - & \underline{8.85} & 15.04 & 0.39 & - & - & - & \underline{8.98} & \underline{15.16} & \underline{0.39} \\

        \midrule
        % MeshPhys $\bigstar$ ($r_{t}=2$) & 2.05 & 7.23 & 0.92 & 13.85 & 23.72 & \underline{0.39} & 7.58 & 18.63 & 0.73 & 19.98 & 28.12 & 0.16 \\

        % MeshPhys $\bigstar$ ($r_{t}=1$) & 3.03 & 8.26 & 0.90 & 10.79 & 18.22 & 0.36 & \textbf{0.46} & \textbf{1.28} & \textbf{1.00} & \underline{9.99} & 17.09 & \underline{0.44} \\

        % \rowcolor{gray!25}  MeshPhys $\bigstar$ & \textbf{0.15} & \textbf{0.25} & \textbf{1.00} & \textbf{8.26} & \underline{16.25} & \textbf{0.51} & \textbf{0.32} & \textbf{0.92} & \textbf{1.00} & \textbf{7.83} & \underline{15.66} & \textbf{0.56} \\

        \rowcolor{gray!10} MeshPhys $\bigstar$ & \textbf{0.18} & \textbf{0.28} & \textbf{1.00} & \textbf{8.18} & 16.03 & \textbf{0.51} & \textbf{0.29} & \textbf{0.86} & \textbf{1.00} & \textbf{7.38} & \textbf{14.70} & \textbf{0.56} \\

        \bottomrule
        % \multicolumn{14}{p{1.00\linewidth}}{
        %     \footnotesize
        %     $\blacktriangle$ Signal Processing; 
        %     $\blacklozenge$ STMap-based Deep Learning;
        %     $\blacksquare$ Video-based Deep Learning; 
        %     $\bigstar$ STGraph-based Deep Learning
        % }
        
    \end{tabular}

    \label{tab:experiments-cross}
\end{table*}

%% file: sec/4d_ablation_stgraph.tex
\subsection{Ablation Study on STGraph}
\label{sec:experiments-ablation-stgraph}
We conduct a systematic ablation of the STGraph design to quantify the contribution of its definition and construction on model performance.
All ablations follow intra-dataset evaluation protocol on MMPD.

\noindent\textbf{Impact of STGraph $(\mathbf{X},\mathbf{A})$ vs.\ STMap $(\mathbf{X})$}.
To isolate the contribution of surface-constrained spatial modeling, we compare the STGraph representation, where node features $\mathbf{X}$ are processed under adjacency $\mathbf{A}$, to an STMap variant using identical node features.
To enable this comparison, we replace the graph convolution operations (GCN) in MeshPhys with 1D convolutions (CNN) over the spatial dimension of the node features, mirroring STMap-based methods.
This disregards surface-based adjacency and treats node features as an unstructured vector, effectively breaking spatial locality and mixing signals across non-contiguous facial regions.
As evidenced in Table~\ref{tab:ablations-stgraph_vs_stmap}, removing surface-constrained feature propagation leads to a clear degradation in performance and robustness (MAE increase of 22.0\% and RMSE increase of 21.1\%).
This highlights a key limitation of STMap-based approaches: while they encode local appearance, they discard surface relationships.
Overall, these results confirm that enforcing spatially localized, surface-constrained feature propagation through $\mathbf{A}$ benefits robust rPPG estimation.

\begin{table}
    \centering
    \caption{Impact of STGraph ($\mathbf{X}$,$\mathbf{A}$) vs. STMap ($\mathbf{X}$) inputs to demonstrate the importance of surface-aligned feature propagation.}
    % \footnotesize
    \begin{tabular}{l c c c c c}
        \toprule

        \multirow{2}{*}{Method} & \multirow{2}{*}{Input} & Conv & MAE $\downarrow$ & RMSE $\downarrow$ & \multirow{2}{*}{\textit{r} $\uparrow$} \\
        & & Op. & (BPM) & (BPM) & \\

        \midrule

        \rowcolor{gray!10} MeshPhys $\bigstar$ & $(\mathbf{X}, \mathbf{A})$ & GCN & \textbf{2.77} & \textbf{7.50} & \textbf{0.85} \\

        MeshPhys $\blacklozenge$ & $(\mathbf{X})$ & CNN & 3.38 & 9.08 & 0.79 \\

        \bottomrule
        
    \end{tabular}

    \label{tab:ablations-stgraph_vs_stmap}
\end{table}

\begin{table}
    \centering
    \caption{Impact of node region definitions to demonstrate the importance of dynamic surface-aligned 3D-aware $\mathcal{M}_{3D}$ representation.}
    % \footnotesize
    \begin{tabular}{l c c c c c c}
        \toprule 

        Node & Per & Surface & 3D & MAE $\downarrow$ & RMSE $\downarrow$ & \multirow{2}{*}{\textit{r} $\uparrow$} \\
        Regions & Frame & Aligned & Aware & (BPM) & (BPM) & \\

        \midrule

        \rowcolor{gray!10} 3D Meshes & \cmark & \cmark & \cmark & \textbf{2.77} & \textbf{7.50} & \textbf{0.85} \\

        2D Meshes & \cmark & \cmark & \xmark &  4.41 & 9.77 & 0.77 \\

        2D Boxes & \cmark & \xmark & \xmark & 4.59 & 10.85 & 0.70 \\

        2D Box & \xmark & \xmark & \xmark & 5.71 & 12.43 & 0.62 \\

        Video & \xmark & \xmark & \xmark & 6.83 & 14.26 & 0.54 \\

        \bottomrule
        
    \end{tabular}

    \label{tab:ablations-noderegions}
\end{table}

\begin{figure}
    \centering
    \small
    \begin{tikzpicture}[node distance=0.50cm and 0.50cm]
        % \fill [fill=blue, opacity=0.10] (0,0) rectangle (\linewidth,0.25\linewidth);

        \node[anchor=south west,inner sep=0] at (0,0){\includegraphics[width=1.00\linewidth]{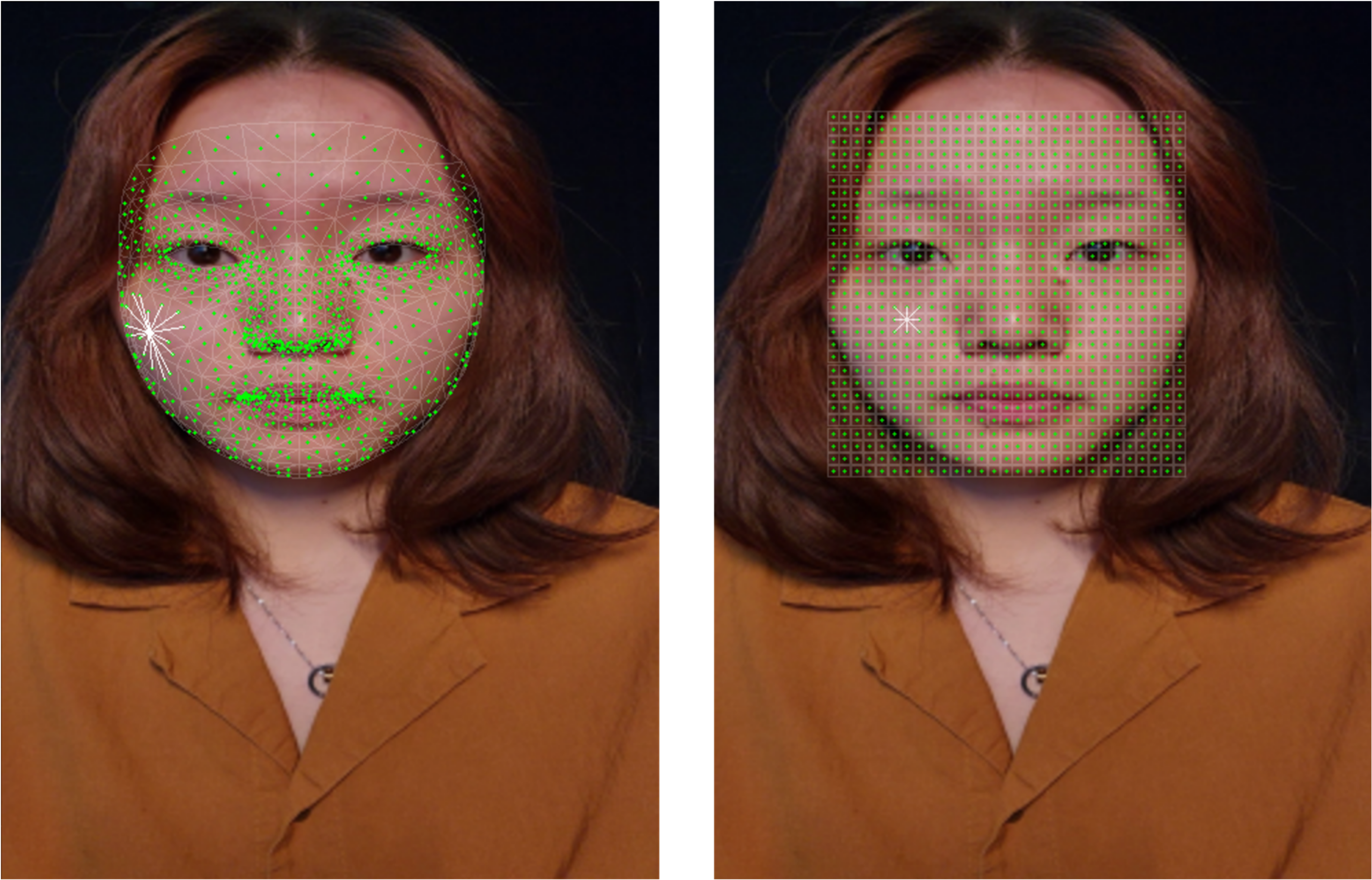}};
        
    \end{tikzpicture}
    \caption{Visualization of STGraph constructions from (left) 3D mesh-based facial landmarks and (right) landmark-derived bounding-box constrained grid.}
    % \DA{THIS FIGURE IS NOT MENTIONED IN THE TEXT}}
    \label{fig:surface-representation}
\end{figure}

\noindent\textbf{Impact of Surface-Aligned 3D-Aware Nodes $(\mathcal{M}_{3D})$}.
To assess the role of surface-aligned and 3D-aware node regions $n_{t,i}$, we ablate five variants with progressively weaker spatial priors.
Specifically, we construct STGraphs using five node region definitions:
(i) \textit{3D Meshes} (default), using 852 triangular faces from MediaPipe~\cite{lugaresi2019mediapipe} with back-face culling to produce surface-aligned, temporally stable, 3D-aware nodes;
(ii) \textit{2D Meshes}, projecting the 3D meshes to the image plane, removing 3D-awareness while retaining region identity and topological consistency;
(iii) \textit{2D Boxes}, a $29 \times 29$ grid placed inside a bounding box derived per frame from the min/max coordinates of the 2D mesh, discarding mesh structure and introducing surface misalignment under motion;
(iv) \textit{2D Box}, using the same $29 \times 29$ grid inside a fixed bounding box from the first frame, further reducing surface consistency; and
(v) \textit{Video}, applying the same grid across the entire image, eliminating facial localization.
Grid-based variants use 8-neighbor connectivity with self-loops for their adjacency (Fig.~\ref{fig:surface-representation}).
All variants use identical region averaging for node feature computation, isolating spatial support as the key factor.
As shown in Table~\ref{tab:ablations-noderegions}, performance degrades consistently as spatial priors are weakened.
Removing 3D-awareness (2D meshes) substantially degrades performance (MAE increase of 59.2\%) due to visual distortion from back-facing areas.
Removing surface alignment (2D boxes) introduces motion-induced noise, further degrading performance (MAE increase of 65.7\%).
Static boxes exacerbate this by freezing node positions over time (MAE increase of 106.1\%).
Performance collapses under a full-frame grid (MAE increase of 146.6\%), which lacks meaningful spatial constraints.
These results confirm that surface-aligned, 3D-aware nodes provide a strong prior, preserving locality, maintaining correspondence under motion, and suppressing irrelevant signals, thereby enabling more robust rPPG estimation.
They also highlight a core limitation of conventional STMap node definitions, whose spatial support fails to provide surface-aligned, 3D-aware features.

\noindent\textbf{Impact of Face-Averaged Node Features $(\mathbf{X})$}.
To evaluate the effectiveness of our node feature definition, we compare face-averaged features—computed by averaging pixels $\mathcal{P}_{t,i}$ over the projection of each 3D mesh face $f_{t,i}$—against alternative image-plane strategies: single-pixel sampling and fixed-size $(H \times W)$ patches centered on the projected face centroid of $f_{t,i}$.
Table~\ref{tab:ablation-nodefeature} shows that using a single pixel results in substantially degraded performance (MAE increase of 48.0\%), highlighting the sensitivity to noise at such a small spatial support.
Enlarging the spatial window improves performance: $3 \times 3$ and $7 \times 7$ patches reduce MAE, with $7 \times 7$ achieving the best quantitative results (MAE reduction of 33.7\% relative to $1 \times 1$).
However, the footprint of fixed-size patches varies across frame sizes (for example, $15 \times 15$ covers 0.29\% of a frame in MMPD but only 0.11\% in PURE), and patches do not adapt to in-plane or out-of-plane motion, resulting in inconsistent sampling and a trade-off between noise sensitivity and spatial support.
In contrast, our mesh-based definition anchors spatial support directly to the facial surface via the 3D mesh.
This yields surface-aligned feature regions that are inherently resolution-invariant, higher-resolution frames provide denser sampling over the projected face.
These findings show that while carefully tuned image-plane patches may perform well under specific conditions, our face-projected node features offer a more stable and resolution-agnostic approach.

\begin{table}
    \centering
    \caption{Impact of node feature construction to demonstrate the importance of resolution-invariant spatial support for appearance encoding. We report average number of pixels per node across the MMPD dataset.}
    % \footnotesize
    \begin{tabular}{l c c c c c c}
        \toprule
        Node Region & Avg. No. Pixels & MAE $\downarrow$ & RMSE $\downarrow$ & \multirow{2}{*}{\textit{r} $\uparrow$} \\
        Averaged & Per Node ($\overline{|\mathcal{P}|}$) & (BPM) & (BPM) & \\

        \midrule
        
        Pixel (1$\times$1) & 1 & 4.10 & 9.08 & 0.79 \\
        
        Region (3$\times$3) & 9 & 2.91 & 8.27 & 0.83 \\
        
        Region (7$\times$7) & 49 &  \textbf{2.72} & \textbf{7.30} & \textbf{0.86} \\
        
        Region (15$\times$15) & 225 & 3.32 & 8.67 & 0.81 \\
        
        % Region (21$\times$21) & 441 &  - & - & - \\

        \midrule
        
        \rowcolor{gray!10} Mesh Face ($f_{t,i}$) & 17.6 & 2.77 & 7.50 & 0.85 \\

        \bottomrule
        
    \end{tabular}

    \label{tab:ablation-nodefeature}
\end{table}

\noindent\textbf{Impact of Edge Connectivity $(\mathcal{E})$}.
To further assess the role of feature propagation, we ablate the edge connectivity $\mathcal{E}$ that defines the adjacency matrix $\mathbf{A}$.
We construct STGraphs using five schemes:
(i) \textit{Shared-Vertex}, connecting mesh faces that share any vertex;
(ii) \textit{Shared-Edge}, connecting only faces that share a full edge;
(iii) \textit{Self-Only}, with no spatial edges;
(iv) \textit{Random}, with edges assigned randomly at the same average degree as (i); and
(v) \textit{Fully-Connected}, linking all nodes.
As shown in Table~\ref{tab:ablation-edgeconnectivity}, shared-vertex connectivity yields the strongest performance, reflecting its ability to support dense, structured propagation across local facial regions.
Removing spatial feature propagation (Self-Only) causes performance to collapse (MAE increase of 534.7\%), confirming that rPPG cannot be modeled exclusively as temporal signals and instead requires spatial interaction across surface regions.
The sparser Shared-Edge variant degrades accuracy (MAE increase of 23.5\%) despite maintaining spatial alignment, suggesting that insufficient local context harms noise rejection.
Random global connectivity with a matched degree performs only slightly worse than Shared-Vertex (MAE increase of 7.6\%), implying that spatial locality—rather than degree alone—drives robustness to region-dependent noise.
Fully-Connected graphs also under-perform (MAE increase of 79.1\%), consistent with loss of locality and signal dilution, where global aggregation obscures subtle spatiotemporal patterns.
Overall, these results emphasize that dense, localized surface connectivity—which enables local spatiotemporal modeling—is a critical inductive bias for rPPG estimation.

\begin{table}
    \centering
    \caption{Impact of edge connectivity $\mathcal{E}$ to demonstrate the importance of dense local connectivity. We report average number of edges per node for each scheme.}
    % \footnotesize
    \begin{tabular}{l c c c c c c c}
        \toprule
        Edge & Avg. No. & MAE $\downarrow$ & RMSE $\downarrow$ & \multirow{2}{*}{\textit{r} $\uparrow$} \\
        Connectivity & Edges & (BPM) & (BPM) & \\

        \midrule

        \rowcolor{gray!10} Shared-Vertex & 12.7 & \textbf{2.77} & \textbf{7.50} & \textbf{0.85} \\

        Shared-Edge & 3.9 & 3.42 & 8.42 & 0.82 \\

        Self-Only & 1.0 & 17.58 & 21.72 & -0.04 \\

        Fully-Connected & 852.0 & 4.96 & 10.72 & 0.73 \\
    
        % Random & 13.0 & 2.98 & 7.96 & 0.84 \\

        \bottomrule
        
    \end{tabular}

    \label{tab:ablation-edgeconnectivity}
\end{table}

\noindent\textbf{Impact of Fine-grained Spatial Features $(|\mathcal{N}|)$}.
To assess the role of spatial granularity, we ablate the number of facial regions $|\mathcal{N}|$ used in the STGraph.
Our default configuration uses $|\mathcal{N}|{=}852$ mesh faces, providing the highest spatial resolution.
Each reduction in $|\mathcal{N}|$ corresponds to spatial coarsening of the STGraph via input pooling, averaging more pixels per node and increasing smoothing.
As shown in Table~\ref{tab:ablation-spatialnodes}, performance degrades consistently as spatial resolution decreases, with $|\mathcal{N}|{=}1$ reducing the model to temporal-only processing and yielding the worst results across all metrics (MAE increase of 271.8\%).
Intermediate resolutions (e.g., $|\mathcal{N}|{=}213$ nodes) offer a balanced compromise, preserving most of the performance benefits with reduced compute cost (MAE increase of only 5.05\% in exchange for an 84.7\% reduction in GMACs), providing flexibility in deployment.
Overall, these results highlight the critical role of spatial resolution and the limitations of purely temporal or overly coarse spatial regions in physiological signal estimation.

\begin{table}
    \centering
    \caption{Impact of number of spatial regions $|\mathcal{N}|$ to demonstrate the importance fine-grained spatial resolution. Computational cost is reported under the default model input size of $\mathbf{X} \in \mathbb{R}^{256 \times 852 \times 3}$.}
    %\small
    \begin{tabular}{l c c c c}
        \toprule        
        
        No. Nodes & MAE $\downarrow$ & RMSE $\downarrow$ & \multirow{2}{*}{\textit{r} $\uparrow$} & Compute $\downarrow$ \\
        ($|\mathcal{N}|$) & (BPM) & (BPM) & & (GMACs)\\ 

        \midrule
        % \midrule

        \rowcolor{gray!10} 852 & \textbf{2.77} & \textbf{7.50} & \textbf{0.85} & 19.14 \\

        % 426 & 2.91 & 7.51 & 0.85 & 7.09 \\

        213 & 2.91 & 7.88 & 0.84 & 2.93  \\

        % 106 & 4.32 & 9.96 & 0.75 & 1.25 \\

        53 & 4.88 & 10.69 & 0.71 & 0.62 \\

        % 26 & 5.20 & 10.53 & 0.73 & 0.29 \\

        13 & 5.26 & 11.87 & 0.62 & 0.19 \\

        % 6 & 6.07 & 12.04 & 0.64 & 0.14 \\

        1 & 10.30 & 16.73 & 0.49 & \textbf{0.12} \\

        \bottomrule
    \end{tabular}

    \label{tab:ablation-spatialnodes}
\end{table}

%% file: sec/4e_ablation_meshphys.tex
\subsection{Ablation Study on MeshPhys \& Losses}
\label{sec:experiments-ablation-meshphys}
We ablate MeshPhys and the loss functions to isolate their impact on performance and demonstrate the benefits of the design.
All ablations follow the intra-dataset evaluation protocol on MMPD unless otherwise stated.

\noindent\textbf{Impact of Multi-kernel Temporal Convolution}.
To assess the importance of adaptive temporal modeling, we compare single-kernel, unweighted multi-kernel, and weighted multi-kernel configurations using depthwise-separable temporal convolutions with varying receptive fields.
As shown in Tables~\ref{tab:ablation-multikernel-intra} and~\ref{tab:ablation-multikernel-cross}, single-kernel models exhibit dataset-specific behavior.
Short kernels ($k{=}3$) retain high-frequency detail and generalize more effectively (PURE$\rightarrow$MMPD: MAE reduction of 36.5\% vs.\ $k{=}9$) but perform poorly in-domain (MMPD: MAE increase of 90.2\% vs.\ $k{=}9$), reflecting their limited capacity to suppress low-frequency distortion and motion artifacts.
Conversely, long kernels ($k{=}9$) provide stronger smoothing and improve in-domain performance (MMPD: MAE reduction of 47.4\% vs.\ $k{=}3$) but degrade under cross-dataset transfer (PURE$\rightarrow$MMPD: MAE increase of 57.5\% vs.\ $k{=}3$), suggesting over-attenuation of temporal variation when applied to less noisy inputs.
Mid-scale kernels ($k{=}5$) yield intermediate performance in both cases.
These results highlight a resolution bottleneck in fixed-kernel models and confirm that temporal filtering needs vary substantially with dataset noise and motion artifacts.

Multi-kernel design aims to address the limitations of fixed temporal filtering by aggregating outputs across branches with different receptive fields.
Uniform branch summation achieves comparable in-domain performance (MMPD: MAE increase of only 3.5\% vs.\ $k{=}9$) but only moderately improves cross-dataset performance (PURE$\rightarrow$MMPD: MAE reduction of 13.4\% vs.\ $k{=}9$).
Branch mixing without weighting offers marginal in-domain gains (MMPD: MAE reduction of 1.1\% vs.\ Sum) but degrades under domain shift (PURE$\rightarrow$MMPD: MAE increase of 12.5\%), suggesting that the mixing overfits to source-domain characteristics.
Weighted branch mixing improves in-domain performance (MMPD: MAE reduction of 19.9\% vs.\ $k{=}9$) and remains competitive under cross-dataset transfer (PURE$\rightarrow$MMPD: MAE increase of 8.1\% vs.\ $k{=}3$), offering a more balanced robustness–generalization trade-off.
These results demonstrate the importance of content-dependent weighting, enabling the model to modulate filtering dynamically based on signal quality, emphasizing detail in clean inputs and applying stronger smoothing in noisy ones.

This module introduces a modest overhead compared to the most balanced single-kernel configuration (parameters +0.24M and compute +5.02 GMACs), as shown in Table~\ref{tab:ablation-multikernel-compute}.
Furthermore, depthwise-separable convolution—which isolates temporal modeling from channel mixing—is more efficient than the full-convolution variant (parameters reduced by 52.5\% and compute by 35.7\%) while also improving performance.

\begin{table}
    \centering
    \caption{Impact of multi-kernel temporal convolution block on MMPD intra-dataset evaluation to demonstrate the importance of adaptive temporal modeling across noise characteristics.}
    % \footnotesize
    \begin{tabular}{l l l l c c c c}
        \toprule

        DWSep & Kernel & Branch & Branch & MAE $\downarrow$ & RMSE $\downarrow$ & \textit{r} $\uparrow$ \\
        
        Conv & Sizes & Weights & Mixing & (BPM) & (BPM) & \\

        \midrule

        % Multi-Kernel Structure with single K
        \cmark & [3] & - & - & 6.58 & 15.23 & 0.58 \\

        \cmark & [5] & - & - & 4.38 & 9.95 & 0.77 \\

        \cmark & [9] & - & - & 3.46 & 8.93 & 0.81 \\

        \midrule
        
        % Multi-Kernel Structure with multi K
        \cmark & [3,5,9] & $1$ & Sum & 3.58 & 8.96 & 0.79 \\

        \cmark & [3,5,9] & $1$ & $\text{Conv}$ & 3.54 & 8.67 & 0.80 \\

        \midrule

        \cellcolor{gray!10}\cmark & \cellcolor{gray!10}[3,5,9] & \cellcolor{gray!10}$\alpha_{p}$ & \cellcolor{gray!10}$\text{Conv}$ & \cellcolor{gray!10}\textbf{2.77} & \cellcolor{gray!10}\textbf{7.50} & \cellcolor{gray!10}\textbf{0.85} \\ %& \cellcolor{gray!10}0.47 & \cellcolor{gray!10}19.17 \\

        \xmark & [3,5,9] & $\alpha_{p}$ & Conv & 4.23 & 9.35 & 0.79 \\

        \bottomrule
        
    \end{tabular}

    \label{tab:ablation-multikernel-intra}
\end{table}

\begin{table}
    \centering
    \caption{Impact of multi-kernel temporal convolution block on cross-dataset PURE$\rightarrow$MMPD evaluation to demonstrate the importance of adaptive temporal modeling across noise characteristics.}
    % \footnotesize
    \begin{tabular}{l l l l c c c c}
        \toprule

        DWSep & Kernel & Branch & Branch & MAE $\downarrow$ & RMSE $\downarrow$ & \textit{r} $\uparrow$ \\
        
        Conv & Sizes & Weights & Mixing & (BPM) & (BPM) & \\

        \midrule

        % Multi-Kernel Structure with single K
        \cmark & [3] & - & - & \textbf{7.57} & \textbf{14.77} & \textbf{0.57} \\ %& 0.19 & 13.36 \\

        \cmark & [5] & - & - & 9.52 & 18.12 & 0.51 \\ %& 0.19 & 13.39 \\

        \cmark & [9] & - & - & 11.92 & 20.64 & 0.29 \\ %& 0.19 & 13.46 \\

        \midrule
        
        % Multi-Kernel Structure with multi K
        \cmark & [3,5,9] & $1$ & Sum & 10.32 & 18.97 & 0.36 \\ % & 0.35 & 16.77 \\

        \cmark & [3,5,9] & $1$ & $\text{Conv}$ & 11.61 & 20.31 & 0.31 \\ %& 0.47 & 19.12 \\

        \midrule

        \cellcolor{gray!10}\cmark & \cellcolor{gray!10}[3,5,9] & \cellcolor{gray!10}$\alpha_{p}$ & \cellcolor{gray!10}$\text{Conv}$ & \cellcolor{gray!10}8.18 & \cellcolor{gray!10}16.03 & \cellcolor{gray!10}0.51 \\

        \xmark & [3,5,9] & $\alpha_{p}$ & Conv & 8.66 & 17.12 & 0.46 \\

        \bottomrule
        
    \end{tabular}

    \label{tab:ablation-multikernel-cross}
\end{table}

\begin{table}
    \centering
    \caption{Computational cost of multi-kernel temporal convolution block reported under the default model input size of $\mathbf{X} \in \mathbb{R}^{256 \times 852 \times 3}$.}% \DA{Table X not mentioned in text}}
    % \footnotesize
    \begin{tabular}{l l l l c c c}
        \toprule

        DWSep & Kernel & Branch & Branch & Parameters $\downarrow$ & Compute $\downarrow$ \\
        
        Conv & Size & Weights & Mixing & (M) & (GMACs) \\

        \midrule

        % Multi-Kernel Structure with single K
        \cmark & [3] & - & - & \textbf{0.23} & \textbf{14.12} \\ %& 0.19 & 13.36 \\

        \cmark & [5] & - & - & 0.23 & 14.16 \\ %& 0.19 & 13.39 \\

        \cmark & [9] & - & - & 0.23 & 14.22 \\ %& 0.19 & 13.46 \\

        \midrule
        
        % Multi-Kernel Structure with multi K
        \cmark & [3,5,9] & $1$ & Sum & 0.35 & 16.77 \\ % & 0.35 & 16.77 \\

        \cmark & [3,5,9] & $1$ & $\text{Conv}$ & 0.46 & 19.09 \\ %& 0.47 & 19.12 \\

        \midrule

        \cellcolor{gray!10}\cmark & \cellcolor{gray!10}[3,5,9] & \cellcolor{gray!10}$\alpha_{p}$ & \cellcolor{gray!10}$\text{Conv}$ & \cellcolor{gray!10}0.47 & \cellcolor{gray!10}19.14 \\ %& \cellcolor{gray!10}0.47 & \cellcolor{gray!10}19.17 \\

        \xmark & [3,5,9] & $\alpha_{p}$ & Conv & 0.99 & 29.78  \\

        \bottomrule
        
    \end{tabular}

    \label{tab:ablation-multikernel-compute}
\end{table}

\noindent\textbf{Impact of Spatial Pooling}.
To evaluate the effect of progressive spatial averaging, we ablate spatial pooling across different coarsening ratios.
As shown in Table~\ref{tab:ablation-spatialpooling}, pooling with a ratio of $r_{s}{=}4$ yields the best accuracy (MAE reduction of 32.9\% vs.\ no pooling), improving over both minimal ($r_{s}{=}2$) and aggressive ($r_{s}{=}8$) reductions.
This suggests that moderate pooling coarsens the graph’s spatial resolution just enough to suppress redundant or noisy activations while retaining localized structure necessary for performance.
It also reduces compute (88.6\% reduction in GMACs) compared to the unpooled baseline, supporting spatial pooling as an effective architectural component for lightweight modeling.

\begin{table}
    \centering
    \caption{Impact of spatial graph pooling ($\mathbf{X}$,$\mathbf{A}$) to demonstrate the importance of hierarchical spatial smoothing of node features.}
    % \footnotesize
    \begin{tabular}{l l c c c c}
        \toprule
        Pooling & Pooling & MAE $\downarrow$ & RMSE $\downarrow$ & \multirow{2}{*}{\textit{r} $\uparrow$} & Compute $\downarrow$ \\
        Op. & Ratio ($r_{s}$) & (BPM) & (BPM) & & (GMACs) \\

        \midrule

        \multirow{4}{*}{AVG} & 8 & 3.53 & 8.41 & 0.82 & \textbf{14.57} \\

        & \cellcolor{gray!10}4 & \cellcolor{gray!10}\textbf{2.77} & \cellcolor{gray!10}\textbf{7.50} & \cellcolor{gray!10}{\textbf{0.85}} & \cellcolor{gray!10}19.14 \\

        & 2 & 3.58 & 8.45 & 0.82 & 38.42 \\

        & - & 4.13 & 9.10 & 0.79 & 168.03 \\

        % \midrule

        % \multirow{4}{*}{MAX} & 8 & - & - & - \\

        % & 4 & - & - & - \\

        % & 2 & - & - & - \\

        % & - & - & - & - \\

        \bottomrule
        
    \end{tabular}

    \label{tab:ablation-spatialpooling}
\end{table}

\noindent\textbf{Impact of Morphological Losses}.
To evaluate the relevance of morphological consistency beyond global correlation objectives, we ablate the derivative-based correlation terms in the loss.
As shown in Table~\ref{tab:ablation-losses-morphology}, eliminating both first- and second-order terms amplifies the degradation (MAE increase of 25.3\%), demonstrating that global correlation alone fails to penalize temporal smoothing or harmonic distortion.
Removing only the second-order correlation loss increases error (MAE increase of 6.5\%), indicating its role in preserving waveform curvature.
Overall, these results show that derivative-based constraints are critical for maintaining waveform morphology and concentrating spectral energy around the physiological fundamental frequency.

\begin{table}[!t]
    \centering
    \caption{Impact of morphological losses on intra-dataset MMPD evaluation to demonstrate the importance of waveform shape supervision.}
    % \footnotesize
    \begin{tabular}{l l l c c c}
        \toprule

        \multirow{2}{*}{$\mathcal{L}_{\rho}^{\varphi}$} & \multirow{2}{*}{$\mathcal{L}_{\Delta\rho}^{\varphi}$} & \multirow{2}{*}{$\mathcal{L}_{\Delta^{2}\rho}^{\varphi}$} & MAE $\downarrow$ & RMSE $\downarrow$ & \multirow{2}{*}{\textit{r} $\uparrow$} \\

        & & & (BPM) & (BPM) & \\

        \midrule

        \rowcolor{gray!10} \cmark & \cmark & \cmark & \textbf{2.77} & \textbf{7.50} & \textbf{0.85} \\

        \cmark & \cmark & \xmark & 2.95 & 7.73 & 0.84 \\

        \cmark & \xmark & \xmark & 3.47 & 8.12 & 0.83 \\

        \bottomrule
        
    \end{tabular}

    \label{tab:ablation-losses-morphology}
\end{table}

\noindent\textbf{Impact of Phase-Shift Alignment}.
To assess the effect of sample-wise misalignment between predicted and ground-truth waveforms, we ablate the phase-shift alignment module that marginalizes the loss over temporal offsets.
As shown in Tab.~\ref{tab:ablation-losses-phaseshift}, performance collapses when disabling this component ($\Phi{=}\pm0$) on poorly synchronized datasets (VIPL-HR: MAE +328.04\%).
While having minimal impact on well-synchronized datasets (MMPD: MAE +0.4\%), we observe slower convergence during training, consistent with unmodeled temporal offsets acting as label noise and inducing conflicting gradients.
Overall, introducing shift tolerance yields more stable supervision and allows the model to prioritize learning accurate waveform morphology over strict sample-wise alignment, which is beneficial under variable capture delays.

\begin{table}
    \centering
    \caption{Impact of phase-shift alignment on intra-dataset evaluation across datasets to demonstrate the importance of consistent time-domain supervision.}
    % \footnotesize
    \begin{tabular}{l l l c c c}
        \toprule

        Evaluation & Phase-Shift & Temp. & MAE $\downarrow$ & RMSE $\downarrow$ & \multirow{2}{*}{\textit{r} $\uparrow$} \\

        Scenario & ($\Phi$) & ($\tau$) & (BPM) & (BPM) & \\

        \midrule

        \multirow{2}{*}{MMPD} & $\pm 50$ & 10.0 & \textbf{2.77} & 7.50 & 0.85 \\

        & $\pm 0$ & - & 2.78 & \textbf{7.37} & \textbf{0.86} \\

        \midrule

        \multirow{2}{*}{VIPL-HR} & $\pm 50$ & 10.0 & \textbf{3.53} & \textbf{6.82} & \textbf{0.84} \\

        & $\pm 0$ & - & 15.11 & 20.69 & 0.25 \\

        \bottomrule
        
    \end{tabular}

    \label{tab:ablation-losses-phaseshift}
\end{table}

\noindent\textbf{Impact of Signal-to-Noise Ratio Weighting}.
To evaluate the effect of SNR-based weighting, we ablate the mechanism that scales the loss according to the spectral quality of each ground-truth signal, thereby attenuating the contribution of unreliable targets.
As shown in Tab.~\ref{tab:ablation-losses-snrweight}, disabling this component ($\gamma{=}0.0$) substantially degrades performance (MAE +71.5\%), indicating that treating high- and low-quality targets uniformly limits the learning of reliable features during training.
These results demonstrate that reliability-aware supervision allows the model to leverage noisy data without being dominated by it, and is critical for learning from real-world datasets with heterogeneous signal quality.

\begin{table}
    \centering
    \caption{Impact of signal-to-noise ratio weighting on intra-dataset MMPD evaluation to demonstrate the importance of attenuating label noise.}
    % \footnotesize
    \begin{tabular}{l l l c c c}
        \toprule

        Weight & Weight & MAE $\downarrow$ & RMSE $\downarrow$ & \multirow{2}{*}{\textit{r} $\uparrow$} \\

        Sharpness ($\gamma$) & Floor & (BPM) & (BPM) & \\

        \midrule

        \rowcolor{gray!10} 0.5 & 0.1 & \textbf{2.77} & \textbf{7.50} & \textbf{0.85} \\

        0.0 & 0.1 & 4.75 & 10.38 & 0.73 \\

        \bottomrule
        
    \end{tabular}

    \label{tab:ablation-losses-snrweight}
\end{table}

\noindent\textbf{Impact of Graph Smoothness}.
To assess the impact of penalizing high-frequency variation across neighboring nodes within the physiological band through the graph smoothness loss $\mathcal{L}_{GS}^{\mathcal{B}}$, we ablate this term from the overall loss.
As shown in Tab.~\ref{tab:ablation-losses-graphsmoothness}, removing this term modestly improves in-domain performance (MMPD: MAE +0.0\% and \textit{r} +2.4\%) but significantly degrades generalization (PURE$\rightarrow$MMPD: MAE +16.7\% and \textit{r} -15.7\%), suggesting it plays a stabilizing role in generalization.
By encouraging smoother node-wise features, the regularizer biases the model towards learning globally coherent features consistent with physiological priors.

\begin{table}
    \centering
    \caption{Impact of graph smoothness regularization on intra- and cross-dataset evaluation to demonstrate the benefits of encouraging globally coherent activations.}
    % \footnotesize
    \begin{tabular}{l l l l c c c}
        \toprule

        Evaluation & \multirow{2}{*}{$\mathcal{L}_{\text{GS}}^{\mathcal{B}}$} & Layer & No. Nodes & MAE $\downarrow$ & RMSE $\downarrow$ & \multirow{2}{*}{\textit{r} $\uparrow$} \\

        Scenario & & ($l$) & ($|\mathcal{N}^{(l)}|$) & (BPM) & (BPM) & \\

        \midrule

        \multirow{2}{*}{MMPD} & \cellcolor{gray!10}\cmark & \cellcolor{gray!10}4 & \cellcolor{gray!10}13 & \cellcolor{gray!10}\textbf{2.77} & \cellcolor{gray!10}7.50 & \cellcolor{gray!10}0.85 \\

        & \xmark & - & - & \textbf{2.77} & \textbf{7.22} & \textbf{0.87} \\

        \midrule

        PURE$\rightarrow$ & \cellcolor{gray!10}\cmark & \cellcolor{gray!10}4 & \cellcolor{gray!10}13 & \cellcolor{gray!10}\textbf{8.18} & \cellcolor{gray!10}\textbf{16.03} & \cellcolor{gray!10}\textbf{0.51} \\

        MMPD & \xmark & - & - & 9.55 & 17.85 & 0.43 \\

        \bottomrule
        
    \end{tabular}

    \label{tab:ablation-losses-graphsmoothness}
\end{table}

%% file: sec/5_discussion.tex
\section{Discussion}
\label{sec:discussion}

\input{sec/5a_compute}
\input{sec/5b_interp}
\input{sec/5c_limitations}
\input{sec/5d_future}

%% file: sec/5a_compute.tex
\subsection{Computational Cost}
\label{sec:discussion-compute}
We report the model and end-to-end computational cost of MeshPhys under the main experimental setups (Secs.~\ref{sec:experiments-intra},~\ref{sec:experiments-cross}) to contextualize its accuracy--efficiency trade-off against video- and STMap-based baselines.

\noindent\textbf{Model Compute}.
At the model level, MeshPhys achieves lower MAE than representative video- and STMap-based models with substantially fewer parameters and moderate compute.
On VIPL-HR, it outperforms RhythmFormer with a 21.7\% MAE reduction, using 85.9\% fewer parameters, 69.4\% less compute, and 56.4\% lower peak memory (Table~\ref{tab:discussion-compute-model}).
Against NEST, it achieves a 25.8\% MAE reduction with 96.6\% fewer parameters, at the cost of 6.7$\times$ higher compute and 56.0\% more memory, due to dense graph operations.
Under domain shift (UBFC$\rightarrow$PURE), it further achieves a 95.2\% MAE reduction relative to NEST, highlighting strong generalization within the SSDG setting.
These results show that MeshPhys reduces model complexity by relying on a structured, surface-aligned STGraph representation, which shifts computational effort from network scale to input encoding.
By explicitly capturing facial topology, the STGraph constrains learning to signal-relevant regions, reducing the burden of suppressing background, motion, and lighting variations—challenges that video-based models must address implicitly.
Overall, MeshPhys achieves high accuracy with a compact model, concentrating its moderate compute budget on structured input representation rather than parameter count.

\noindent\textbf{Pipeline Compute}.
End-to-end, MeshPhys remains computationally competitive even when full pipeline costs—including preprocessing—are considered.
As shown in Table~\ref{tab:discussion-compute-endtoend}, the complete pipeline (face detection, mesh extraction, STGraph construction, and model inference) achieves a 64.9\% reduction in parameter count, 42.4\% lower compute, and 30.7\% lower peak VRAM usage compared to the model-only cost of RhythmFormer in our setup.
The pipeline comprises lightweight, modular stages, but its throughput is primarily limited by STGraph construction, which accounts for 86.8\% of inference time.
This cost profile reflects a design trade-off: rather than learning spatial structure implicitly through large model capacity, MeshPhys encodes it explicitly via the STGraph.
This enables a smaller model to focus computation on signal-relevant regions, maintaining a practical end-to-end budget while supporting strong performance.

\noindent\textbf{Contextualizing Preprocessing Costs}.
Most STMap-based methods and the majority of video-based approaches rely on dedicated preprocessing steps such as face detection and feature extraction.
In many cases, these stages are implemented as separate, fixed components external to the model. 
For example, RhythmFormer and NEST use once-off face detection, minimizing runtime cost but reducing robustness to motion or pose changes. 
In contrast, methods like BeatFormer and Dual-TL apply per-frame detection and masking to handle dynamic scenes, increasing computational overhead.
MeshPhys follows the latter pattern, using per-frame detection and feature extraction, but uniquely integrates surface-aligned features and topology through STGraph construction. 
This representation-centric design enables strong generalization with a compact model, and the associated preprocessing cost is integral to the method rather than auxiliary to it.

\begin{table}
    \centering
    \caption{Model-only computational cost under model-specific spatial input sizes. 
    Results are obtained on an NVIDIA GeForce RTX 3090 using FP32 with temporal length $|\mathcal{T}|=256$ sequence and averaged across 10 runs with a batch size of 1.
    Best and second best results are formatted as \textbf{bold} and \underline{underline} respectively.}
    % \footnotesize
    \begin{tabular}{l c c c c c}
        \toprule
        \multirow{3}{*}{Method} & 
        Params & Compute & Throughput & Peak \\
        & $\downarrow$ & $\downarrow$ & $\uparrow$ & VRAM $\downarrow$ \\
        & (M) & (GMACs) & (kFPS) & (GiB) \\

        \midrule

        NEST $\blacklozenge$ & 13.98 & \textbf{2.48} & \textbf{34.19} & \textbf{0.50} \\

        rPPG-MAE $\blacklozenge$ & 88.30 & \underline{4.96} & 13.70 & 1.42 \\

        \midrule

        DeepPhys $\blacksquare$ & 2.23 & 58.21 & 11.21 & 1.77 \\

        PhysNet $\blacksquare$ & \underline{0.77} & 35.23 & \underline{20.95} & \underline{0.61} \\

        PhysFormer $\blacksquare$ & 7.43 & 83.06 & 5.99 & 2.12 \\

        EfficientPhys $\blacksquare$ & 2.16 & 28.75 & 15.53 & 1.26 \\

        RhythmFormer $\blacksquare$ & 3.33 & 62.55 & 10.75 & 1.79 \\

        RhythmMamba $\blacksquare$ & 5.14 & - & 0.83 & 1.46 \\

        \midrule

         MeshPhys $\bigstar$ & \textbf{0.47} & 19.14 & 11.93 & 0.78 \\

        \bottomrule
        
    \end{tabular}

    \label{tab:discussion-compute-model}
\end{table}

\begin{table}
    \centering
    \caption{End-to-end computational cost of STGraph construction and MeshPhys inference.
    Results are obtained on an NVIDIA GeForce RTX 3090 using FP32 with temporal length $|\mathcal{T}|=256$ sequence and averaged across 10 runs with a batch size of 1.}
    %\footnotesize
    \begin{tabular}{l c c c c c}
        \toprule
        \multirow{3}{*}{Method} & 
        \multirow{2}{*}{Params. $\downarrow$} & \multirow{2}{*}{Compute $\downarrow$} & \multirow{2}{*}{Throughput $\uparrow$} & Peak \\
        & & & & VRAM $\downarrow$ \\
        & (M) & (GMACs) & (kFPS) & (GiB) \\

        \midrule

        BlazeFace $\blacksquare$ & 0.10 & 7.87 & 29.91 & 1.17 \\

        FaceMesh $\blacksquare$ & 0.60 & 8.95 & 35.73 & 1.10 \\
        
        % $\mathcal{G}_{f}$ construction & - & 0.01 & 0.26 & 0.31 \\

        $\mathcal{G}_{f}$ construction & - & 0.01 & 0.26 & 0.31 \\

        MeshPhys $\bigstar$ & 0.47 & 19.14 & 11.93 & 0.78 \\

        % \midrule
        \midrule

        End-to-end & 1.17 & 35.97 & 0.25 & 1.17 \\

        \bottomrule
        
    \end{tabular}

    \label{tab:discussion-compute-endtoend}
\end{table}

%% file: sec/5b_interp.tex
\subsection{Interpretability}
\begin{figure}
    \centering
    \small
    \begin{tikzpicture}
        % \fill [fill=blue, opacity=0.10] (0,0) rectangle (\linewidth,0.25\linewidth);
% 
        \node[anchor=south west,inner sep=0] at (0,0){\includegraphics[width=1.00\linewidth]{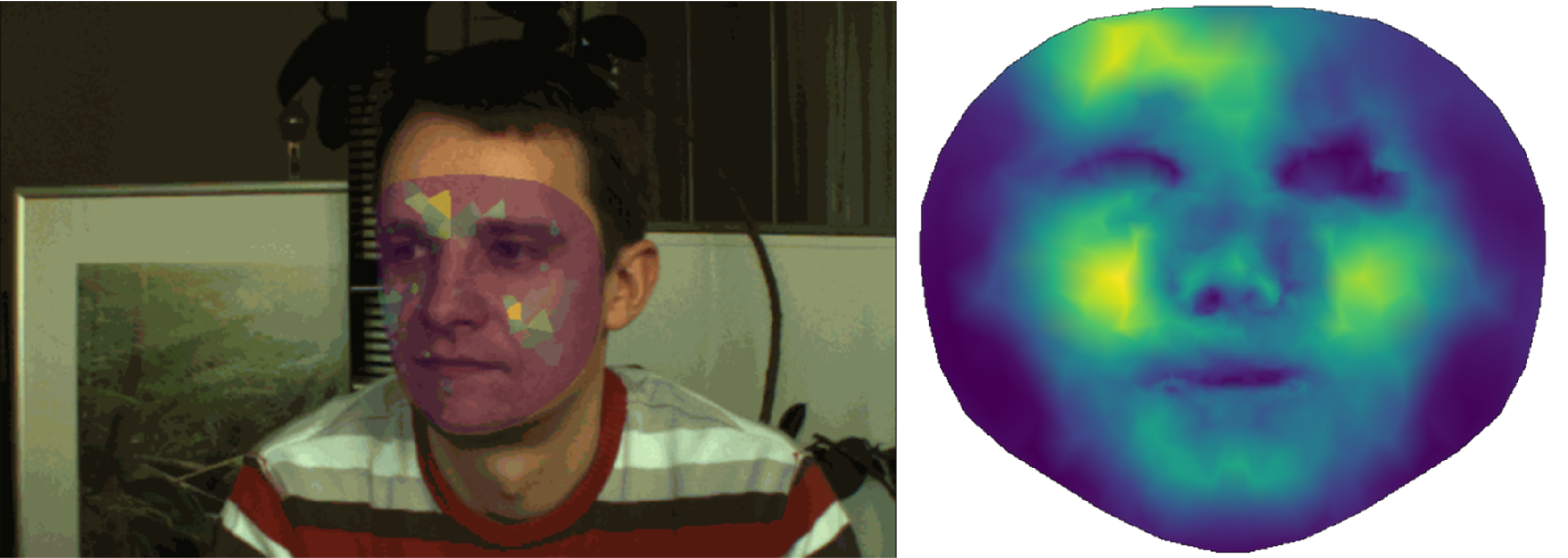}};
        
    \end{tikzpicture}
    \caption{Visualization of gradient-weighted activations from MeshPhys with respect to the 3D facial mesh within a video frame (left) and the canonical 2D facial mesh (right).}
    \label{fig:discussion-interpretability}
\end{figure}

MeshPhys operates directly on a structured representation of the facial surface via the STGraph, resulting in predictions are inherently tied to specific regions over time.
The dependence on 3D landmarks serves a dual role: it filters out frames with unreliable access to the facial surface and provides an anchor for interpreting model activations. 
Fig.~\ref{fig:discussion-interpretability} visualizes the average gradient-weighted activation from the output of the multi-kernel temporal convolution block in the second layer, projected both in image space (left) and interpolated across canonical coordinate space (right). 
High-importance regions consistently localize to known pulse-rich areas—such as the cheeks, forehead, and sides of the nose—aligning with findings from prior physiological studies~\cite{kim2021assessment} and from empirical observations of existing methods~\cite{chen2018deepphys,stphys2024}.
This alignment between learned activations and physiologically relevant regions reinforces the model’s interpretability and suggests that its performance stems not from overfitting to superficial cues, but from leveraging underlying physiological signal structure.
As such, interpretability in MeshPhys is a direct outcome of its explicit surface-aware design.

%% file: sec/5c_limitations.tex
\subsection{Limitations}
MeshPhys inherits practical limitations and inherent from its explicit reliance on the facial surface.
Firstly, STGraph construction depends on accurate 3D landmark detection to generate node features $\mathbf{X}$, and requires contiguous detections to form valid training sequences—though missing frames could be masked at inference.
This acts as an interpretable flag for unsuitable frames, however it remains a side-effect.
As shown in Table~\ref{tab:discussion-datasets}, this issue is not present in PURE and UBFC-rPPG, is minimal in VIPL-HR, and affects one video significantly in MMPD (\textit{p9\_19}).
Furthermore, inaccurate landmarks can distort the facial mesh, producing spatial misalignment that entangles rigid motion and non-rigid motion changes into the node features $\mathbf{X}$.
More fundamentally, MeshPhys assumes consistent node correspondence across frames, limiting its use to 3D surface representations with topologically consistent meshes across frames. 
While these constraints reduce general applicability, they enable stronger inductive bias and more interpretable, physiologically grounded representations.

\begin{table}[!t]
    \centering
    \caption{Frames without detected landmarks per dataset.}
    %\footnotesize
    \begin{tabular}{l c c c}
        \toprule

        \multirow{2}{*}{Dataset} & No. Missing & No. Total & \% Missing \\
        & Frames & Frames & Frames \\

        \midrule

        PURE & 0 & 123,276 & 0.00\% \\
        UBFC-rPPG & 0 & 81,401 & 0.00\% \\
        MMPD & 1,896 & 1,188,000 & 0.16\% \\
        VIPL-HR & 1,687 & 1,856,138 & 0.09\% \\ 

        \bottomrule
        
    \end{tabular}

    \label{tab:discussion-datasets}
\end{table}

%% file: sec/5d_future.tex
\subsection{Future Work and Broader Impact}
There are numerous opportunities to expand STGraph-based physiological modeling. 
While $\mathcal{G}_f$ captures facial surface topology and color through the node and edge definitions, alternative graph formulations—redefining $\mathcal{V}$, $\mathcal{E}$, $\mathbf{X}$, and $\mathbf{A}$—could incorporate learned connectivity, non-spatial edge relations, or richer node features.
Furthermore, relaxing the fixed topology assumption of the STGraph would allow for dynamic or non-canonical meshes, supporting the use of more general 3D facial surface representations while preserving the spatial grounding that underpins interpretability.
Beyond such structural extensions, more expressive graph architectures may better capture and extract physiological variation upon the facial surface. 
MeshPhys serves as a simple yet effective baseline, but there remains significant headroom in terms of model capacity and use within alternative learning strategies.

Within a broader context, robustness is critical for physiological monitoring in real-world settings—such as telehealth, consumer health tracking, and low-resource environments—where controlled capture conditions or specialized sensors are often unavailable.
MeshPhys addresses these constraints by using the STGraph to enforce surface-aligned spatial structure, enabling accurate and stable rPPG estimation from monocular RGB input alone.
The spatial grounding of the STGraph further introduces interpretability by design, allowing attribution to localized facial regions and enabling frame-level quality assessment through mesh visibility and stability.
Together, these properties position STGraph-based models as a promising foundation for deployable physiological systems that are both reliable and transparent.

%% file: sec/6_conclusion.tex
\section{Conclusion}
\label{sec:conclusion}
This work demonstrates that constraining spatiotemporal modeling to the facial surface serves as a principled and strong prior for rPPG estimation.
The STGraph encodes both appearance and structure through surface-aligned, 3D-aware, fine-grained node features $\mathbf{X}$ and spatial adjacency $\mathbf{A}$, grounded on a fixed-topology facial mesh.
Operating on this representation, MeshPhys achieves state-of-the-art or competitive in-domain accuracy and strong cross-dataset generalization, using only a lightweight model.
By explicitly aligning the model's receptive field with the facial surface, this approach enables coherent, surface-constrained feature propagation that supports robust physiological signal estimation.
Ablation studies confirm the benefit of these design choices and reveal clear advantages over STMap-based methods that discard structural priors.
Overall, these findings establish STGraphs as a powerful and general modeling framework for remote physiological measurement, with MeshPhys serving as an efficient and effective baseline for future research.